\documentclass[mnsc,nonblindrev]{informs3} 

\OneAndAHalfSpacedXI 

\usepackage{endnotes}
\let\footnote=\endnote

    \usepackage[colorlinks=true,hypertexnames=false,bookmarks=false,urlcolor=blue, citecolor=blue,linkcolor=blue,bookmarksopen=false,draft=false]{hyperref}

\usepackage{tikz}
\usetikzlibrary{shapes.geometric, arrows.meta, positioning}
\usepackage{amsmath}
\usepackage{bm} 
\usepackage{framed}
\usepackage{longtable}
\usepackage{array}
\usepackage{booktabs}

\newcolumntype{P}[1]{>{\raggedright\arraybackslash}p{#1}}
\newcolumntype{N}[1]{>{\centering\arraybackslash}p{#1}}
\usepackage[outdir=./]{epstopdf}

\usepackage{graphicx}
\usepackage{multirow}
\usepackage{booktabs} 

\usepackage{siunitx}
\usepackage{microtype}

\usepackage{listings}
\lstdefinestyle{promptstyle}{
  basicstyle=\footnotesize\ttfamily,
  breaklines=true,
  breakatwhitespace=false,
  columns=fullflexible,
  keepspaces=true,
  frame=single,
  framerule=0.3pt,
  xleftmargin=0.5em,
  xrightmargin=0.5em,
  aboveskip=0.75\baselineskip,
  belowskip=0.75\baselineskip,
  showstringspaces=false
}

\usepackage[ruled,vlined,noend]{algorithm2e} 
\SetKwInput{KwInit}{Initialize}
\SetKwInput{KwIn}{Receive}
\SetKwInput{KwOut}{Return}
\SetAlFnt{\small       }

\SetAlCapFnt{\small       }
\SetAlCapNameFnt{\small       }
\usepackage{algorithmic}
\algsetup{linenosize=\small       }

\usepackage{url}

\newcolumntype{C}[1]{>{\centering\arraybackslash}p{#1}}

\usepackage{natbib}
 \bibpunct[, ]{(}{)}{,}{a}{}{,}%
 \def\bibfont{\small}%
\TheoremsNumberedThrough      
\ECRepeatTheorems

\EquationsNumberedThrough    

\usepackage{amsbsy, amsfonts, amsgen, amsmath, amsopn, amssymb, amstext,
    amsxtra, bezier, color, enumerate, graphicx, latexsym, verbatim,
    pictexwd, supertabular, url, dsfont,leftidx, mathrsfs, appendix, 
    setspace,scalerel,mathtools,arydshln,stmaryrd,upgreek,bbm}
\allowdisplaybreaks

\makeatletter
\newcommand{\raisemath}[1]{\mathpalette{\raisem@th{#1}}}
\newcommand{\raisem@th}[3]{\raisebox{#1}{$#2#3$}}
\makeatother

\newcommand{\algo}{\textsc{AIPS}}

\newcommand{\Inv}{\ensuremath{\bm{{I}}}}
\newcommand{\InvOn}{\ensuremath{{I}^\mathrm{on}}}
\newcommand{\InvPipe}{\ensuremath{{I}^\mathrm{pipe}}}
\begin{document}



\RUNAUTHOR{Author Name et al.}

\RUNTITLE{Short Title Here}

\TITLE{Automated Design of Inventory Policy with\\ Large Language Models: An Exploratory Study}

\ARTICLEAUTHORS{%
\AUTHOR{Fenghua Yang, Preet Baxi}
\AFF{University of Michigan, Ann Arbor, MI 48109, USA (\EMAIL{yfenghua@umich.edu}, \EMAIL{preetb@umich.edu})}
\AUTHOR{Yi Zhang}
\AFF{Stanford University, Stanford, CA 94305, USA (\EMAIL{yz5195@stanford.edu})}
\AUTHOR{Stefanus Jasin}
\AFF{University of Michigan, Ann Arbor, MI 48109, USA (\EMAIL{sjasin@umich.edu})}
\AUTHOR{Yanzhe Lei}
\AFF{Queen's University, Kingston, ON K7L 3N6, Canada (\EMAIL{yl64@queensu.ca})}
\AUTHOR{Mo Liu}
\AFF{University of North Carolina, Chapel Hill, NC 27599, USA (\EMAIL{mo\_liu@unc.edu})}
\AUTHOR{Parshan Pakiman}
\AFF{University at Buffalo, State University of New York, Buffalo, NY 14260, USA  (\EMAIL{parshanp@buffalo.edu})}
%

}

\ABSTRACT{%
    Firms making inventory decisions increasingly have access to rich operational data, advanced optimization tools, and rapidly improving large language models (LLMs). Typically, data characterize the operating environment, optimization selects parameter values within a prespecified inventory policy class, and LLMs support tasks such as coding and decision analysis. We develop an integrated framework that combines these resources to automate inventory policy design. Given demand data, the framework iteratively uses an LLM to generate parameterized policy classes and an external solver to optimize its parameters within each class. Across 30 lost-sales inventory instances, the mean cost reduction relative to optimized base-stock benchmarks increases from 17.5\% after one generation to 30.0\% after ten generations. 
    Parameter optimization is central to this performance: an LLM-only variant performs substantially worse, whereas optimization-guided feedback improves policy quality, accelerates search, and directs the LLM toward better policy classes rather than merely better parameter values within a fixed class. The strongest discovered policies are also interpretable: they combine recognizable inventory-control motifs, including capped orders, discounted or weighted pipeline inventory, and threshold-based replenishment logic. The search thereby produces new policy-class functional forms that, to our knowledge, have not previously been studied in the lost-sales inventory literature. These functional forms are not specified ex ante but emerge from the search process. 
    Moreover, after their parameters are re-optimized, three discovered policy classes achieve average cost reductions of 21.75\% -- 22.60\% across 10,064 new inventory instances. Overall, the results show that data-driven parameter optimization can guide LLM-based search over a broad space of inventory policy classes and identify high-performing, interpretable, and transferable decision rules.
}

\KEYWORDS{Large Language Models; inventory control; automated policy design;  evolutionary search}

\maketitle

\vspace{-16pt}
\section{Introduction}
    
    Optimization models have long formed the backbone of inventory decision making, providing principled methods for characterizing, computing, and improving replenishment policies within specified policy classes \citep{arrow1951optimal,scarf1960optimality,veinott1965computing,zipkin2000foundations,porteus2002foundations}. The growth of operational data has enabled firms to complement these models with richer information, including sales histories, inventory records, and contextual features, to improve demand forecasting, demand estimation, and replenishment decisions \citep{iyer1992analysis,fildes2009effective,dehoratius2008retail,huh2011adaptive,ding2024feature}. More recently, large language models (LLMs) and generative artificial intelligence have emerged as a third resource, with growing potential to support optimization practice, supply-chain management, and broader managerial decision making \citep{jiang2024survey,chen2025optichat,csaszar2024artificial,cohen2026supply,mckinsey2025supplychain}. Thus, an important question is not simply whether LLMs will be used in practice, but how they can be used effectively. In inventory control, these three resources are typically assigned distinct roles: data support forecasting and demand estimation, optimization improves decisions within prespecified policy classes, and LLMs assist with tasks such as modeling, coding, and decision analysis.\looseness=-1
    
    Can data, optimization, and LLMs instead be integrated into a systematic mechanism for designing effective replenishment policies? This question is important because inventory policy design involves two distinct choices. The first is the \textit{policy class}: the functional form of the decision rule that maps the available inventory-state information to an ordering action. The second is the \textit{parameter vector} that specifies a particular policy within that class. For example, the base-stock policy class is defined by a single-parameter decision rule, with each value of the base-stock level specifying a particular policy. Data characterize the operating environment, including demand patterns, lead times, service requirements, and cost tradeoffs. Given a policy class, optimization can then identify effective parameter values within that class. Neither data nor parameter optimization alone, however, determines which functional form should define the policy class. In practice, the policy class is often selected heuristically based on computational tractability or ease of interpretation, potentially at the expense of performance. This limitation motivates a broader question: \textit{can LLMs expand the policy-design space beyond prespecified classes and support the automated discovery of previously unexplored policy classes, rather than selecting among or optimizing existing ones?}\looseness=-1
    
    The choice of policy class is particularly consequential in lost-sales inventory systems with positive lead times. In these systems, replenishment decisions must jointly account for on-hand inventory, outstanding pipeline orders, stochastic demand, and the irreversibility of lost sales. Although base-stock policies can perform well in some settings, they need not be optimal when lead times are positive because the value of an order depends on both the amount and timing of inventory already in the pipeline. Prior work has therefore examined alternative policy classes, including constant-order, capped base-stock, and projected-inventory-level policies \citep{zipkin2008old,huh2009asymptotic,xin2021capped,jaarsveld2024projected,liu2026inventory,xin2026capped}. This literature demonstrates that the choice of policy class can materially affect performance. It does not, however, provide a general mechanism for systematically searching beyond policy classes specified by researchers in advance.\looseness=-1
    
    LLMs create a new opportunity for policy-class search. Recent work shows that LLMs can generate executable code, interact with optimization models, and support automated heuristic design \citep{jiang2024survey,chen2025optichat,romera2024funsearch,liu2024evolution,ye2024reevo}. In particular, methods such as FunSearch, Evolution of Heuristics, ReEvo, and AlphaEvolve embed LLMs in evaluation-guided, reflective, or evolutionary search procedures that iteratively generate, evaluate, and refine executable algorithms \citep{romera2024funsearch,liu2024evolution,ye2024reevo,novikov2025alphaevolve}. Concurrent work by \citet{huang2026invevolve} applies LLM-based evolutionary search to generate interpretable inventory policies for online deployment under nonstationary demand, using reinforcement learning to evaluate and improve candidate policies. \cite{baek2026llms} studies a simple single-query protocol in which an LLM generates either an instance-specific solution or a reusable algorithm for well-specified problems in inventory control, queueing-network control, and assortment optimization, and finds that the strongest model is competitive with specialized methods. \citet{baek2026ai} studies complementarities among operations-research heuristics, LLM agents, and human decision makers in multiperiod inventory control; in their setting, the LLM participates in or supports period-by-period ordering decisions rather than searching ex ante over stationary policy classes. Our setting and methodological focus differ from both streams. We study offline policy-class design for stationary lost-sales inventory systems and use off-the-shelf LLMs to generate parameterized policy classes, external numerical optimization to select parameter values within each generated class, and simulation to evaluate the resulting policies. This separation allows us to isolate whether optimization-guided feedback helps the LLM generate better policy classes, rather than merely better parameter values within previously generated classes.\looseness=-1
    
    Our Automated Inventory Policy Search (\algo{}) framework builds on the discovery-oriented approach of recent LLM-based heuristic-design methods and adapts it to the sequential decision-making setting of inventory control. Given an inventory environment defined by demand data and instance-specific primitives, including lead time, holding cost, and lost-sales cost, \algo{} uses an LLM to generate a pool of parameterized policy classes. Each class defines a mapping from the inventory state, including on-hand and pipeline inventory, to an order quantity. These classes may range from standard classes, such as the single-parameter base-stock class and the two-parameter $(s,S)$ class, to richer classes with larger and more flexible parameter spaces. For each generated class, numerical optimization selects parameter values by minimizing a simulation-based empirical cost objective. The resulting optimized policy is then evaluated, and the best-performing policies, together with their class-defining code, optimized parameters, and performance summaries, are returned to the LLM to guide the generation of the next pool of candidate classes. Therefore, \algo{} alternates between two interconnected steps: \emph{policy-class search}, in which the LLM generates candidate policy classes, and \emph{parameter optimization}, in which numerical optimization identifies effective parameter values within each generated class. Simulation-based evaluation closes the loop by measuring the performance of the optimized policies and feeding this information, together with the optimized parameter values, back to the LLM to guide subsequent policy-class search.\looseness=-1
     
    We conduct an extensive computational study of \algo{} in lost-sales inventory systems with positive lead times. Rather than attempting to characterize optimal policies, we use a sequence of experiments to evaluate how effectively LLMs can generate useful classes of inventory policies and how the framework's components contribute to performance. First, benchmark comparisons assess whether \algo{} discovers policy classes containing policies that outperform optimized base-stock policies. Second, optimization ablations quantify the contribution of numerical parameter optimization and examine whether feeding optimized parameter values and policy performance back to the LLM improves subsequent policy-class search. Third, structural analyses identify the inventory-control motifs that recur in the best-performing discovered policies. Fourth, generalization experiments vary lead times, lost-sales-penalty-to-holding-cost ratios, and demand distributions to evaluate whether discovered policy classes remain effective beyond the instances on which they were developed. Fifth, cross-model comparisons measure the sensitivity of the framework's performance to the choice of LLM backbone. Collectively, these experiments evaluate both the quality of the policies discovered by \algo{} and the mechanisms through which optimization-guided feedback improves the search over policy classes.

\subsection{Contributions and Related Literature}

Our work contributes to several streams of research: lost-sales inventory control, data-driven inventory decision making, LLM-based automatic heuristic design, and approximate policy search. We organize our contributions around these connections.

    \begin{itemize}

        \item \textbf{Automated policy search for inventory control.} We develop a computational framework that expands inventory policy design beyond a prespecified menu of standard heuristics. The framework uses an off-the-shelf LLM to generate executable parameterized policy classes and numerical optimization to identify effective policies within those classes; the resulting optimized policies and performance information then guide subsequent structural search. Recent LLM-based automatic heuristic-design methods use evaluator-guided or evolutionary procedures to generate and improve executable algorithms \citep{romera2024funsearch,liu2024evolution,ye2024reevo,novikov2025alphaevolve}. Concurrent inventory studies pursue related but distinct objectives: \citet{huang2026invevolve} develop an end-to-end framework that uses a reinforcement-learning-trained LLM to evolve interpretable inventory policies with statistical safety guarantees for deployment, whereas \citet{baek2026ai} study complementarities among LLM agents, operations-research heuristics, and human decision makers in sequential inventory decision making. 
        
        In contrast, we study the ex ante discovery of reusable policy classes for stationary lost-sales inventory systems. Our main contribution is to show how data and optimization can guide a general-purpose LLM to search a broad space of executable inventory decision rules and uncover effective policy classes that need not be specified by the researcher in advance. More importantly, because both LLM and optimization components are modular and rely on off-the-shelf technologies, our framework can directly benefit from advances in LLM reasoning and code generation as well as improvements in the efficiency and capability of commercial optimization solvers, without requiring changes to its underlying architecture.\looseness=-1

        \item \textbf{Exploratory computational study.} We provide systematic computational evidence on when LLM-guided inventory policy search creates value, why it does so, and how robust that value is across problem settings and LLM backbones. Our study varies key primitives of lost-sales inventory systems, including lead times from two to eight periods, multiple lost-sales-penalty-to-holding-cost ratios, and demand environments spanning bounded continuous, bounded discrete, light- to moderate-tailed unbounded, and heavy-tailed or highly skewed distributions. We also vary central features of the search procedure, most importantly whether LLM-generated policy classes undergo numerical parameter optimization, and compare performance across six LLM backbones: GPT-5 Nano, Gemini 2.5 Flash-Lite, Grok 4.1 Fast Non-Reasoning, GPT-5 Mini, DeepSeek V3, and Gemini 3 Flash. Existing studies of LLM-based heuristic design and LLM-enabled inventory decision making demonstrate that LLMs can generate executable decision rules and support sequential decisions, but offer limited evidence on their performance across heterogeneous inventory environments and algorithmic configurations. Our experiments address this gap by identifying when LLM-generated policy classes outperform optimized base-stock benchmarks, quantifying the contribution of numerical parameter optimization and optimization-guided feedback, characterizing recurring structural motifs in the discovered classes, evaluating their transfer across inventory environments, and measuring the sensitivity of the framework to the choice of LLM backbone.\looseness=-1

        \item \textbf{Optimization-guided evaluation for LLM-based policy search.} We show that numerical optimization should be an integral part of the evaluation mechanism in LLM-guided policy search. Rather than evaluating the initial policy proposed by the LLM, we evaluate the best policy that can be obtained within each generated policy class by optimizing its parameters. Recent LLM-based heuristic-design methods use evaluators to execute candidate programs and guide reflective or evolutionary search \citep{romera2024funsearch,liu2024evolution,ye2024reevo,novikov2025alphaevolve}. We extend this paradigm by introducing an optimization-guided evaluator that measures the quality of a policy class through its optimized representative rather than its initial parameterization. Computationally, this distinction proves important: optimization not only improves the performance of individual policies but also provides substantially more informative feedback, directing the LLM toward increasingly effective policy classes across generations.

        \item \textbf{Structurally meaningful discovered policies.} We show that \algo{} discovers policy classes with interpretable and economically meaningful functional form rather than arbitrary black-box decision rules. Across problem instances, the strongest classes repeatedly combine recognizable inventory-control motifs, including order caps, discounted or weighted representations of pipeline inventory, and threshold-based replenishment logic. These motifs relate naturally to established lost-sales policies: capped base-stock policies augment base-stock logic with order caps, while projected-inventory-level policies adjust replenishment decisions using anticipated future inventory positions \citep{xin2021capped,jaarsveld2024projected,xin2026capped}. Our findings extend this literature by showing that such motifs need not be specified by the researcher ex ante but can emerge from LLM-guided search and be recombined into richer parameterized policy classes. More broadly, \algo{} can be viewed as an evolutionary search over inventory-specific basis-function representations of replenishment policies: the LLM proposes transformations and combinations of inventory-state variables that define candidate policy classes, while numerical optimization selects their associated parameters. In this respect, our work contributes to recent research on automating approximation design for Markov decision processes \citep{pakiman2025self,nadarajah2025self,adelman2025dynamic,lagzi2026using} by using LLMs to discover interpretable structural representations tailored to inventory control.

        \item \textbf{Strong and transferable discovered policy classes.} We show that \algo{} can discover replenishment-policy classes containing policies that frequently outperform optimized base-stock benchmarks and remain effective beyond the instances on which they are generated. This comparison is consequential because base-stock policies are natural and widely used benchmarks, yet they can perform poorly in lost-sales systems with positive lead times. Prior work has therefore developed alternative policy classes, including constant-order, capped base-stock, and projected-inventory-level policies \citep{zipkin2008old,huh2009asymptotic,xin2021capped,jaarsveld2024projected}. 
        
        Rather than manually designing a new heuristic policy class for lost-sales inventory problems, we develop a mechanism that automatically generates policy classes with diverse structures, optimizes policies within those classes, and retains the strongest classes through simulation-based evaluation. We further assess whether the discovered classes capture reusable replenishment structures rather than instance-specific rules by transferring them across changes in lead time, cost ratio, and the demand distribution's family, mean, and variability, with their parameters reoptimized for each new environment. This analysis goes beyond conventional out-of-sample evaluation on new demand realizations from a fixed environment and extends robustness questions in the lost-sales literature from prespecified policy classes to classes generated automatically through search. We also examine sensitivity to the LLM backbone and find that model choice affects the economic performance of the discovered policies, while numerical parameter optimization reduces, but does not eliminate, this variation. Collectively, these results indicate that the value of LLMs in inventory control lies not in making period-by-period ordering decisions directly, but in expanding the set of effective and transferable policy classes available for optimization and evaluation.
    \end{itemize}

\subsection{Organization} 
    The remaining part of the paper is organized as follows. In \S\ref{sec:model}, we introduce the lost-sales inventory model and the benchmark policies. In \S\ref{sec:methodology}, we present the \algo{} framework, including the LLM-based policy-generation step, the numerical optimization step, the simulation-based evaluation step, and the feedback mechanism across generations. In \S\ref{sec:main}, we describe the experimental design, report the main benchmark and ablation results, and analyze the structural motifs that emerge in the discovered policies. In \S\ref{sec:generalization}, we study transferability across lead times, cost ratios, and demand distributions. In \S\ref{sec:external}, we examine the role of optimization. We use DeepSeek (specifically, \texttt{deepseek-chat-v3-0324}) as the default LLM backbone in \S\S\ref{sec:main}--\ref{sec:external}, holding the generator fixed to isolate the value of the proposed data--LLM--optimization framework relative to benchmark policies, its ability to uncover reusable structural motifs, and the role of the external optimizer. In \S~\ref{sec:diff-LLMs}, we relax this fixed-backbone design and evaluate the framework across multiple LLM backbones. We conclude in \S\ref{sec:conclusion}. Additional numerical results are reported in the online supplement. A GitHub repository will be made publicly available to facilitate potential future adaptations or extensions of \algo{}.\looseness=-1

\section{Model and Optimization Objective}\label{sec:model}

    We consider a single-item, periodic-review inventory system over a finite horizon of \(T\) periods with deterministic lead time \(L\). Time is indexed by \(t=1,\ldots,T\). At the beginning of period \(t\), before the order scheduled to arrive in that period is received, the inventory state is \(\Inv_t:=\big(\InvOn_t,\InvPipe_{t,1},\ldots,\InvPipe_{t,L}\big)\). Throughout, we use boldface symbols to denote vectors. Here, \(\InvOn_t\) denotes the on-hand inventory before receipt of the arriving order, and \(\bm{\InvPipe}_t:=\big(\InvPipe_{t,1},\ldots,\InvPipe_{t,L}\big)\) denotes the pipeline-inventory vector. Its component \(\InvPipe_{t,\ell}\) is the quantity scheduled to arrive at the beginning of period \(t+\ell-1\); in particular, \(\InvPipe_{t,1}\) is the quantity arriving in period \(t\). We assume that the initial inventory state \(\Inv_1\) is known to the decision maker and that \(\InvOn_1=0\).\looseness=-1
    
    At the beginning of period \(t\), the first pipeline component arrives, yielding available inventory \(\bar I_t:=\InvOn_t+\InvPipe_{t,1}\). The decision maker then places an order \(a_t\geq 0\), which enters the pipeline and arrives at the beginning of period \(t+L\). An i.i.d. demand realization \(D_t\) is then drawn from distribution \(\tilde D\), and any unmet demand is lost. Consequently, the quantity sold is \(\min\{\bar I_t,D_t\}\), the next-period on-hand inventory is \(\InvOn_{t+1}=(\bar I_t-D_t)^+\), and lost sales are \((D_t-\bar I_t)^+\), where \(x^+:=\max\{x,0\}\). At the end of the period, the remaining pipeline components shift forward by one position, and the newly placed order enters as the final component. The resulting state transition is\looseness=-1
    \[
        \Inv_{t+1}:=\Big((\bar I_t-D_t)^+,\InvPipe_{t,2},\ldots,\InvPipe_{t,L},a_t\Big),
    \qquad
    \bm{\InvPipe}_{t+1}:=\Big(\InvPipe_{t,2},\ldots,\InvPipe_{t,L},a_t\Big).
    \]
    The period-\(t\) cost function is given by $c(\Inv_t,D_t):=h(\bar I_t-D_t)^+ + p(D_t-\bar I_t)^+$,
    where \(h\geq 0\) and \(p\geq 0\) are the unit holding and lost-sales costs, respectively. We normalize the variable ordering cost to zero. Thus, the first term is the holding cost incurred on inventory remaining after demand is served, whereas the second term is the penalty incurred on unmet demand.\looseness=-1
    
    We focus on stationary replenishment policies represented by decision rules of the form \(\pi:\mathcal I\rightarrow\mathcal A\), where \(\mathcal I\subseteq \mathbb{R}_+^{L+1}\) is the inventory-state space and \(\mathcal A:=[0,\bar a]\) is the feasible action space, with \(\bar a<\infty\) denoting an upper bound on the order quantity. In our simulations, we set 
    \(\bar a\) sufficiently large so that the upper bound does not bind. For each state \(\Inv\in\mathcal I\), policy \(\pi\) prescribes an order quantity \(\pi(\Inv)\in\mathcal A\). Stationarity means that the decision rule does not depend explicitly on the period index, while the Markov property means that past demands, states, and actions affect the current decision only through the current inventory state \(\Inv\). Let \(\Pi\) denote the set of feasible stationary Markov replenishment policies. This set includes policies from standard classes, such as the single-parameter base-stock class and the two-parameter \((s,S)\) class, as well as policies from richer parameterized classes with more flexible functional forms and larger parameter spaces. For a policy \(\pi\in\Pi\), let \(\Inv_t^\pi\) denote the inventory state in period \(t\) when policy \(\pi\) is followed. The objective is to solve the policy optimization problem
    \begin{equation}\label{eq:PO}\tag{PO}
        \inf_{\pi\in\Pi}J(\pi),
        \qquad
        J(\pi):=\mathbb{E}_{\tilde D}^{\pi}\!\left[\sum_{t=1}^{T} c\!\left(\Inv_t^{\pi},D_t\right)\right],
    \end{equation}
    where the expectation is taken with respect to the demand distribution \(\tilde D\) and policy \(\pi\).

    Tackling \eqref{eq:PO} poses two fundamental challenges. First, the demand distribution \(\tilde D\) is unknown, so the expectation in the objective function \(J(\pi)\) cannot be evaluated directly. A standard approach is therefore to replace the expectation in \eqref{eq:PO} with an empirical objective constructed from observed demand data, as studied in the data-driven inventory literature (e.g., \citealt{ban2019big,lin2022data,zhang2025sampling}). Second, the unrestricted policy space \(\Pi:=\{\pi:\mathcal I\rightarrow\mathcal A\}\) is an infinite-dimensional function space, making direct optimization over policies generally intractable. This difficulty is compounded by the curse of dimensionality: the inventory state includes both on-hand inventory and the entire pipeline, so its dimension grows with the lead time. Consequently, a broad literature restricts attention to tractable policy classes for lost-sales inventory systems and establishes conditions under which such policies perform near-optimally \citep{zipkin2008old,huh2009asymptotic,goldberg2016asymptotic,xin2021capped,jaarsveld2024projected}.\looseness=-1

    \textbf{Data-driven cost function.} For the first challenge, we approximate the expectation with respect to the unknown demand distribution \(\tilde D\) using sample-average approximation. Let \(\mathcal D:=\big\{\big(\widehat D_1^n,\ldots,\widehat D_T^n\big):n=1,\ldots,N\big\}\) denote a dataset consisting of \(N\geq 1\) demand sample paths, each of length \(T\). We then define the empirical policy optimization problem as
    \begin{equation}\label{eq:EPO}\tag{EPO}
        \inf_{\pi\in\Pi}\widehat J(\pi),
        \qquad
        \widehat J(\pi):=\frac{1}{N}\sum_{n=1}^{N}\sum_{t=1}^{T}c\!\left(\Inv_t^{n,\pi},\widehat D_t^n\right),
    \end{equation}
    where \(\big(\widehat D_1^n,\ldots,\widehat D_T^n\big)\) is the \(n\)th observed demand trajectory and \(\Inv_t^{n,\pi}\) is the inventory state at the beginning of period \(t\) when policy \(\pi\) is simulated along that trajectory. Thus, \(\widehat J(\pi)\) is the sample-average cost of policy \(\pi\) over the training trajectories and provides an empirical approximation of \(J(\pi)\). Because all candidate policies are evaluated on the same demand trajectories, \(\widehat J\) provides a common empirical criterion for policy comparison.

    \textbf{Parametric policy class.} For the second challenge, we approximate the unrestricted policy space \(\Pi\), which contains all feasible mappings from inventory states to ordering decisions, by a parameterized policy class. Specifically, a \emph{policy class} \(\Pi_{\Theta}:=\left\{\pi_{\bm{\theta}}:\bm{\theta}\in\Theta\right\}\subseteq\Pi\) is characterized by a \emph{functional form} \(\pi_{\bm{\theta}}(\cdot):\mathcal I\rightarrow\mathcal A\) and a \emph{feasible parameter space} \(\Theta\subseteq\mathbb{R}^{d}\). The functional form determines how the inventory state enters the ordering rule, while the parameter vector \(\bm{\theta}\in\Theta\) selects a particular policy within the class. Thus, for any fixed \(\bm{\theta}\), \(\pi_{\bm{\theta}}(\cdot)\) is a fully specified policy that maps an inventory state \(\Inv\in\mathcal{I}\) to the order quantity \(\pi_{\bm{\theta}}(\Inv)\in\mathcal{A}\). This representation gives rise to two distinct tasks: policy-class design, which determines the functional form and parameter space, and within-class parameter optimization, which selects parameter values for the inventory environment under consideration. For example, the base-stock policy class, denoted $\Pi_{\Theta}^{\mathrm{BS}}$, is given by the one-dimensional parameter \(\bm{\theta}:=(\theta_{\mathrm B})\) and is defined as
    \[
        \Pi_{\Theta}^{\mathrm{BS}} := \left\{  \pi_{\mathrm B}: \pi_{\mathrm B}(\Inv) = \min\left\{\bar a,\left(\theta_{\mathrm B}-I^{\mathrm{pos}}\right)^+\right\},\ \theta_{\mathrm B}\in\Theta \right\},
    \]
    where \(I^{\mathrm{pos}}:=\InvOn+\sum_{\ell=1}^{L}\InvPipe_{\ell}\) denotes the inventory position associated with state \(\Inv\). Here, \(\bar a\) continues to denote the upper bound on the order quantity. Each value \(\theta_{\mathrm B}\in\Theta\) therefore specifies one policy \(\pi_{\mathrm B}\) within the base-stock class. The best policy within this class is obtained by selecting \(\theta_{\mathrm B}\in\Theta\) to minimize the population cost \(J(\pi_{\mathrm B})\) when the demand distribution \(\tilde D\) is known, or the empirical cost \(\widehat J(\pi_{\mathrm B})\) when only demand dataset \(\big(\widehat D_1^n,\ldots,\widehat D_T^n\big)\) is available.

\section{LLM-Guided Inventory Policy Search} \label{sec:methodology}

\begin{figure}[t]
    \centering
    \caption{Overview of the AIPS procedure.}
    \includegraphics[width=\textwidth]{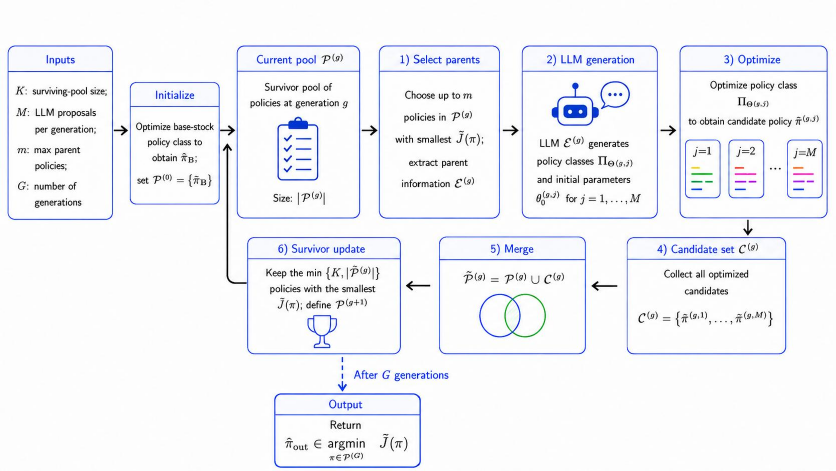}
    \label{fig:aips_workflow}\vspace{-30pt}
\end{figure}

In this section, we present \algo{} that integrates data, optimization, and LLMs. Specifically, given a finite search budget, at each search generation, an LLM proposes policy classes and an external numerical optimizer searches for low-cost parameter values within each class. These components are embedded in an evolutionary loop in which high-performing policies guide subsequent policy-class proposals. We present these components in the following order: \S\ref{sec:class-proposal} introduces policy-class design via the LLM, \S\ref{sec:parameter-optimization} introduces within-class parameter optimization, \S\ref{sec:evolutionary_search} combines the two components into the full evolutionary feedback loop, and \S\ref{sec:class-validation} collects validation and numerical implementation details. Figure~\ref{fig:aips_workflow} provides an overview of the
complete \algo{} procedure.

\subsection{Policy-Class Design via an LLM}
\label{sec:class-proposal}

    Rather than manually prespecifying a policy class (e.g., a base-stock policy), \algo{} uses an LLM to iteratively generate parameterized policy classes. For each generated class, a numerical optimizer searches over its feasible parameter domain to identify a parameter vector that yields low empirical training cost, thereby producing an optimized policy within that class. High-performing optimized policies are then used to guide the generation of new policy classes. We refer to the policies selected for this purpose as \emph{parent policies}; each parent is therefore an instantiated policy from a previously generated class, together with its optimized parameter vector. In \S\ref{sec:evolutionary_search}, we specify how these parent policies are selected. Given one or more parents, the prompt provides the LLM with each parent's functional form, optimized parameter vector, and empirical training cost. Using this information, the LLM is asked to generate exactly one new parameterized policy class, together with an initial parameter vector and a feasible parameter domain.

\begin{figure}[t]
\centering
\caption{Illustrative LLM prompt for generating a new parameterized policy class from parent policies.}
\begin{lstlisting}[style=promptstyle]
PROBLEM INSTANCE STATEMENT
Lead time: L = 6
Holding cost: h = 1
Lost-sales cost: p = 2
Demand data: demand sample paths

TASK DESCRIPTION
Given parent policies, generate a new policy class that achieves lower empirical training cost on the demand data. 
The functional form of the policy class should have the following input and output format:
Input: current on-hand inventory and pipeline inventory
Output: one nonnegative scalar order quantity

PARENT POLICY SET
PARENT POLICY 1
Functional form of the base-stock policy:
    ...
Parameter vector for base-stock policy:
    ...
Empirical training cost:
    ...
PARENT POLICY 2
Functional form of the constant order policy:
    ...
Parameter vector for constant order policy:
    ...
Empirical training cost:
    ...
    
REQUIREMENT
Return exactly one new candidate policy class consisting of:
1. functional form of the policy class by Python code;
2. an initial value for every tunable parameter; and
3. a feasible lower and upper bound for every parameter.

\end{lstlisting}
\label{fig:illustrative_prompt}
\end{figure}

\textbf{Illustrative LLM prompt.} To make the LLM interface concrete, consider a policy-class generation call in which a base-stock policy and a constant-order policy serve as parent policies. Figure~\ref{fig:illustrative_prompt} presents an abridged version of the prompt supplied to the LLM. The prompt contains four top-level blocks. The \texttt{PROBLEM INSTANCE STATEMENT} specifies the inventory instance introduced in \S\ref{sec:model}, including the lead time \(L\), the unit holding and lost-sales costs \(h\) and \(p\), and the demand dataset \(\mathcal D\). The \texttt{TASK DESCRIPTION} defines the generation objective: using the parent policies as guidance, the LLM is asked to propose a new parameterized policy class that, after parameter optimization, can yield a policy with lower empirical cost on the given demand data. The \texttt{PARENT POLICY SET} provides the feedback used to guide the structural search. For each parent policy \(\widehat{\pi}\), the prompt reports its functional form, optimized parameter vector, and corresponding empirical cost. Finally, the \texttt{REQUIREMENT} block specifies the output of a single LLM proposal call. The LLM returns
\(
\left(\Pi_{\Theta},\bm{\theta}_0\right),
\)
where
\(
\Pi_{\Theta}
=
\left\{
\pi_{\bm{\theta}}:\bm{\theta}\in\Theta
\right\}
\)
is the generated policy class, \(\Theta\) is its feasible parameter domain, and \(\bm{\theta}_0\in\Theta\) is the LLM-supplied initial parameter vector. The example omits formatting and validation instructions that do not affect this conceptual input--output relationship. The complete prompt template is reported in \ref{app:prompt_template}.

    The LLM prompt in Figure~\ref{fig:illustrative_prompt} can be formalized by representing the information supplied about the parent policies. Let
    \[
        \mathcal E := \left\{ \left( \Pi_{\Theta_j}, \widehat{\bm\theta}_j, \widehat J\!\big(\pi_{\widehat{\bm\theta}_j}\big) \right) \right\}_{j=1}^{m}
    \]
    denote the parent set. Each element records a previously generated policy class \(\Pi_{\Theta_j}\), the optimized parameter vector \(\widehat{\bm\theta}_j\) that instantiates a parent policy \(\pi_{\widehat{\bm\theta}_j}\) within that class, and the resulting empirical training cost. Thus, \(\mathcal E\) contains both the structural information and the performance feedback shown to the LLM. Given this parent set, one LLM call produces a new parameterized policy class and an initial parameter vector:
    \begin{equation*}
        (\Pi_\Theta,\bm\theta_0)
        \gets
        \texttt{LLM}\!\left(\mathcal E\right),
    \end{equation*}
    where \(\Pi_\Theta\) is the newly proposed policy class and \(\bm\theta_0\in\Theta\) is the LLM-supplied initialization for subsequent parameter optimization. The inventory instance, training data, and prompt instructions are fixed within a search run and are therefore suppressed from the arguments of \(\texttt{LLM}\). Thus, the LLM primarily searches over policy structures rather than parameter values within a fixed structure. It may modify the state transformations, algebraic operations, branching logic, number of tunable parameters, and their roles. The LLM-supplied \(\bm\theta_0\) serves only as an initialization; proposal quality is evaluated after the external optimizer searches for low-cost parameters within \(\Pi_\Theta\).

\subsection{Within-Class Parameter Optimization} \label{sec:parameter-optimization}
    Given an LLM output \( (\Pi_\Theta,\bm\theta_0) \), the optimization  module searches over \(\Theta\) while holding the functional form \(\pi_{\bm\theta}(\cdot)\) fixed. The corresponding
    within-class empirical optimization problem is
    \begin{align}\tag{EPO$(\Pi_\Theta)$}
        \min_{\bm\theta\in\Theta}
        \widehat J
        \left(
            \pi_{\bm\theta}
        \right)
        &=
        \min_{\bm\theta\in\Theta}
        \frac{1}{N}
        \sum_{n=1}^{N}
        \sum_{t=1}^{T}
        c\left(
            \Inv_t^{n,\pi_{\bm\theta}},
            \widehat D_t^n
        \right).
        \label{eq:within-class-optimization}
    \end{align}
    Problem~\eqref{eq:within-class-optimization} is the empirical policy-optimization problem~\eqref{eq:EPO} restricted to the generated class \(\Pi_{\Theta}\). Every parameter vector is evaluated on the same fixed training trajectories, so differences in empirical cost are not caused by demand resampling.

    Because the numerical search is conducted under a finite computational budget, we do not assume that it solves problem ~\eqref{eq:within-class-optimization} to local or global optimality. Instead, among the parameter vectors successfully evaluated by the numerical routine, let \(\widehat{\bm\theta}\) denote one attaining the lowest empirical cost. We define
\begin{equation*}
\widehat\pi
:=
\pi_{\widehat{\bm\theta}}
\end{equation*}
as the retained policy associated with \(\Pi_\Theta\). We summarize the numerical procedure by the function
\begin{equation}\label{eq:para_optimize}
        \widehat\pi
    \gets
    \texttt{Optimize}
    \left(
        \Pi_\Theta,\bm\theta_0
    \right).
\end{equation}
The external optimizer may change only \(\bm\theta\); it cannot alter the policy's
state variables, transformations, algebraic operations, branch
conditions, or any other structural feature encoded by
\(\pi_{\bm\theta}(\cdot)\). Consequently,
\(\texttt{LLM}\) searches across policy classes,
whereas \(\texttt{Optimize}\) searches for a policy
within a fixed class.\looseness=-1

\SetKwInput{KwReceive}{Receive} \SetKwInput{KwInitialize}{Initialize} \SetKwInput{KwReturn}{Return} 
\setlength{\algomargin}{7pt} 
\LinesNumbered 
\begin{algorithm}[h] 
\DontPrintSemicolon 
\SetAlgoLined 
    \KwReceive{Maximum surviving-pool size \(K\); number of LLM proposals per generation \(M\); maximum number of parent policies \(m\); and number of generations \(G\).} 
    \KwInitialize{Optimize the base-stock policy class to obtain \(\widehat\pi_{\mathrm B}\), and set \(\mathcal P^{(0)}\gets\{\widehat\pi_{\mathrm B}\}\).} 
    \For{\(g=0,1,\ldots,G-1\)}{ 
        \(\mathcal C^{(g)}\gets\emptyset\)\;
        \(\mathcal E^{(g)}\gets\) the parent information of the \(\min\{m,|\mathcal P^{(g)}|\}\) policies $\pi\in\mathcal P^{(g)}$ with the smallest  \(\widehat J(\pi)\) \nllabel{alg:select-parents}\;
        \For{\(j=1,\ldots,M\)}{
            Generate \(\big(\Pi_{\Theta^{(g,j)}},\bm\theta_0^{(g,j)}\big)\gets\texttt{LLM}\big(\mathcal E^{(g)}\big)\) and compute \(\widehat\pi^{(g,j)}\gets\texttt{Optimize}\big(\Pi_{\Theta^{(g,j)}},\bm\theta_0^{(g,j)}\big)\)\;
            Set \(\mathcal C^{(g)}\gets\mathcal C^{(g)}\cup\{\widehat\pi^{(g,j)}\}\) \nllabel{alg:generate-offspring}\;
        } \vspace{6pt}
        Define \(\widetilde{\mathcal P}^{(g)}\gets\mathcal P^{(g)}\cup\mathcal C^{(g)}\) \nllabel{alg:merge-policies}\; 
        Update \(\mathcal P^{(g+1)}\) to the \(\min\{K,|\widetilde{\mathcal P}^{(g)}|\}\) policies $\pi\in\widetilde{\mathcal P}^{(g)}$ with the smallest values of \(\widehat J(\pi)\) \nllabel{alg:update-pool}\; 
    } \vspace{6pt}
    \KwReturn{\(\widehat\pi_{\mathrm{out}}\in\operatorname*{arg\,min}_{\pi\in\mathcal P^{(G)}}\widehat J(\pi)\).} 
\caption{\normalfont{Automated Inventory Policy Search (AIPS)}} 
\label{alg:llm_heuristic_design} 
\end{algorithm}

\subsection{Algorithm} \label{sec:evolutionary_search} Algorithm~\ref{alg:llm_heuristic_design} combines the two components introduced above---LLM-based policy-class generation and numerical parameter optimization---within an evolutionary search. The algorithm maintains a surviving pool \(\mathcal P^{(g)}\) of optimized policies at generation \(g\). Starting from an optimized base-stock policy, each generation proceeds sequentially as follows. First, the best-performing policies in the current surviving pool are selected as parents (line~\ref{alg:select-parents}). Second, for each of \(M\) offspring, the LLM uses the parent information to propose a new parameterized policy class, and the external optimizer searches within that class to obtain an optimized policy; the resulting policy is then added to the offspring set (line~\ref{alg:generate-offspring}). Finally, the offspring and incumbent policies are pooled, and the \(K\) policies with the lowest empirical training costs survive to the next generation (lines~\ref{alg:merge-policies}--\ref{alg:update-pool}). After \(G\) generations, the algorithm returns the best policy remaining in the surviving pool. The search is governed by four hyperparameters: \(K\), the maximum size of the surviving pool; \(M\), the number of LLM proposals generated in each generation; \(m\), the maximum number of parent policies supplied to each LLM call; and \(G\), the number of generations. 

    \textbf{Initialization.} The algorithm begins from a standard policy class that provides a valid incumbent before the first LLM call. Throughout the paper, we use the base-stock policy as this initializer. For target level \(\theta_{\mathrm B}\), its ordering rule is $\pi_{\mathrm B}\left(\Inv_t;\theta_{\mathrm B}\right):=\left[\theta_{\mathrm B}-\left(\InvOn_t+\sum_{k=1}^{L}\InvPipe_{t,k}\right)\right]^+$.  We treat this rule as a single-parameter policy class and apply the same numerical optimization module used for LLM-generated classes to determine its parameter. Let \(\widehat\pi_{\mathrm B}\) denote the resulting optimized base-stock policy. The initial surviving pool is therefore $\mathcal P^{(0)}:=\{\widehat\pi_{\mathrm B}\}$.  This initialization ensures that the search starts from a feasible and interpretable policy and that the surviving pool is nonempty when the first generation begins. Other user-specified policy classes could be used as initializers, although doing so may change the subsequent search trajectory. 

    \textbf{Parent selection.} At the beginning of generation \(g\), the algorithm creates an empty offspring set \(\mathcal C^{(g)}\) and selects parent policies from the current surviving pool \(\mathcal P^{(g)}\). Specifically, it selects the \(\min\{m,|\mathcal P^{(g)}|\}\) policies with the smallest empirical training costs. As described in the preceding subsection, the information supplied to the LLM for each selected parent consists of its generating policy class, optimized parameter vector, and empirical training cost. We collect this information in the parent set \(\mathcal E^{(g)}\). Thus, the LLM observes both the structures that have performed well so far and the parameter values and costs obtained after numerical optimization. 

    \textbf{Policy-class generation and parameter optimization.} Holding the parent set \(\mathcal E^{(g)}\) fixed, the algorithm makes \(M\) separate LLM calls. On proposal \(j\), the LLM generates \[ \left(\Pi_{\Theta^{(g,j)}},\bm\theta_0^{(g,j)}\right)\gets\texttt{LLM}\left(\mathcal E^{(g)}\right), \] where \(\Pi_{\Theta^{(g,j)}}\) is a new parameterized policy class and \(\bm\theta_0^{(g,j)}\in\Theta^{(g,j)}\) is its LLM-supplied initialization. The external optimizer then searches over the feasible parameter domain \(\Theta^{(g,j)}\) to obtain an optimized policy \[ \widehat\pi^{(g,j)}\gets\texttt{Optimize}\left(\Pi_{\Theta^{(g,j)}},\bm\theta_0^{(g,j)}\right). \] Importantly, the object retained by the algorithm is this optimized policy, not the policy obtained directly from the LLM-supplied initialization. Each optimized policy is added to the offspring set, giving $\mathcal C^{(g)}:=\{\widehat\pi^{(g,j)}:j=1,\ldots,M\}$. 
    
    \textbf{Survival selection and population update.} After all \(M\) proposals have been optimized, the offspring are combined with the policies that survived from the previous generation: $\widetilde{\mathcal P}^{(g)}:=\mathcal P^{(g)}\cup\mathcal C^{(g)}$. The policies in \(\widetilde{\mathcal P}^{(g)}\) are ranked according to their empirical training costs, and the next surviving pool \(\mathcal P^{(g+1)}\) consists of the \(\min\{K,|\widetilde{\mathcal P}^{(g)}|\}\) lowest-cost policies. Hence, a newly generated policy survives only if its optimized performance is sufficiently competitive with that of the incumbents. Because the surviving policies become the candidates for parent selection in the next generation, better-performing policy structures receive greater opportunity to influence subsequent LLM proposals. \looseness=-1
    
    \textbf{Output.} After \(G\) generations, the algorithm returns the policy with the lowest empirical training cost among the final survivors: $\widehat\pi_{\mathrm{out}}\in\operatorname*{arg\,min}_{\pi\in\mathcal P^{(G)}}\widehat J(\pi)$. Thus, selection throughout the search is based on performance \emph{after} within-class parameter optimization. The external optimizer therefore plays two roles: it calibrates the parameters of each proposed class, and, through the resulting empirical costs, it determines which policy structures are retained and subsequently fed back to the LLM. This separation allows the LLM to focus primarily on searching over policy structures while the numerical optimizer searches for effective parameter values within each proposed structure.

\subsection{Validation and Numerical Implementation Details}
\label{sec:class-validation}
To operationalize \algo{}, the implementation must
also determine whether a generated policy class is admissible and whether a
particular parameter vector can be executed reliably in simulation.
We therefore apply validation at two levels---class-level static
validation and parameter-specific runtime validation---and then
summarize the numerical routine used for within-class search. These
implementation safeguards are important for executability and
reproducibility, but do not alter the conceptual search procedure
described above.

\textbf{Class-level static validation.}
Before numerical search, the implementation parses the generated source code and the annotations identifying optimizable parameters. A proposal is discarded if the code cannot be parsed; if the parameter metadata are missing or internally inconsistent; if \(\bm\theta_0\notin\Theta\); or if the declared parameter domain is invalid. The policy function must satisfy the common interface: it may use only the current on-hand and pipeline-inventory state, must return a scalar order quantity, and may not depend explicitly on the period index, trajectory index, previous function calls, or persistent hidden state. Determinism and stationarity are enforced through hard-coded validation rules. A proposal that fails any of these checks causes \(\texttt{Optimize}\) to return \(\varnothing\) without beginning numerical search.

\textbf{Parameter-specific runtime validation.}
Passing class-level static validation does not imply that every \(\bm\theta\in\Theta\) produces an executable policy. Each parameter vector queried by the numerical routine is therefore checked during simulation. A parameter vector is considered feasible only if every training trajectory completes within the prespecified evaluation time limit and every visited state produces a finite scalar order quantity. Execution errors, timeouts, and nonfinite outputs make that parameter vector invalid. If \(\mathcal V=\varnothing\), \(\texttt{Optimize}\) returns \(\varnothing\), and the proposal does not enter the offspring pool.

\textbf{Numerical implementation.}
In the implementation, we use the \texttt{minimize} routine in \texttt{scipy.optimize} with \texttt{L-BFGS-B}, finite-difference gradient estimates, bounded parameter domains, and $15$ optimizer iterations. The simulation objective can be nonsmooth because of branch conditions, lost-sales dynamics, clipping, type conversions, and integer rounding. We therefore use \texttt{L-BFGS-B} as a standardized bounded numerical search routine, not as a method guaranteed to identify a local or global optimum. Among all valid parameter vectors evaluated under the budget, the routine retains the one with the lowest empirical training cost, as defined in~\eqref{eq:EPO}.\looseness=-1

\section{Performance and Structure of Discovered Policies}\label{sec:main}

This section evaluates the performance and structure of the policies discovered by \algo{}. In \S\ref{sec:implementation_details}, we define the 30 benchmark instances, describe the experimental design, and introduce the performance metrics used to compare \algo{} with the optimized base-stock policy. In \S\ref{sec:performance}, we quantify the cost reductions achieved by \algo{} over generations and examine how these gains vary with lead time, cost parameters, and demand distributions. In \S\ref{sec:motif_analysis}, we characterize recurring structural motifs in the discovered policies by \algo{}. In \S\ref{represent_trajectory}, we trace a representative evolutionary run of \algo{} to illustrate how the incumbent policy evolves across generations and how structural motifs are combined and refined throughout the search.

\subsection{Problem Instances and Performance Metrics} \label{sec:implementation_details}

We consider 30 instances of the lost-sales inventory control problem in \S\ref{sec:model} to benchmark \algo{}. Each instance corresponds to a unique combination of demand distribution, lead time, and cost parameters. Specifically, we consider three lead times \(L\in\{2,4,6\}\), two cost settings \((h,p)\in\{(1,2),(1,5)\}\), and five demand specifications: Poisson demand, Exponential demand, and three Normal demand settings with standard deviations \(10\), \(30\), and \(50\). The mean demand is normalized to 100 in all cases. Thus, the benchmark set spans a range of operating regimes, from short- to long-lead-time systems and from relatively stable to highly variable demand environments.

For each instance \(k\), we run the evolutionary search for \(G=10\) generations. In each generation, the LLM generate \(M=10\) offsprings. We refer to one complete run of the algorithm over all \(G\) generations as a \emph{repeat}. For each instance, we conduct 10 independent repeats using the same training data. In other words, the repeats differ only in the randomness of LLM sampling. For repeat \(r\), let \(\mathcal P_{k,r}^{(g)}\) denote the surviving pool after generation \(g\). We rank the policies in this pool by empirical training cost and define the lowest empirical training cost attained through generation \(g\) in repeat \(r\) as 
\[
\widehat J_{k,r}^{(g)}
:=
\min_{\pi\in\mathcal P_{k,r}^{(g)}}
\widehat J_k(\pi).
\]

For each instance \(k\), we compare the policies discovered by \algo{} with the optimized base-stock policy for the same instance. For each instance \(k\), let \(\widehat J_k(\cdot)\) denote the empirical training cost evaluated on the same training trajectories used by \algo{}. Using the same notation in Algorithm~\ref{alg:llm_heuristic_design}, the empirical cost of the optimized base-stock policy is \(\widehat J_k(\hat{\pi}_B)\), which will be used as the benchmark. We define the percentage cost reduction for instance \(k\) at generation \(g\) in repeat $r$ as
\[
\mathrm{CR}_{k,r}^{(g)}
:=
100
\left(
1-
\frac{\widehat J_{k,r}^{(g)} }{\widehat J_k(\hat{\pi}_B) }
\right).
\]
Thus, \(\mathrm{CR}_{k,r}^{(g)}>0\) indicates that the best policy discovered after $g$ generations has a lower cost than the optimized base-stock cost for instance \(k\), with larger values corresponding to greater cost reductions. In this section, we often report the mean and standard error of $\mathrm{CR}_{k,r}^{(g)}$ across all ten repeats and a family of problem instances with certain features (e.g., all instances with Poisson demand).\looseness=-1

\subsection{Performance Results}\label{sec:performance}
    
We first report some summary statistics of all the generated policies (see Table~\ref{tab:cr-by-instance-generation-compact} in Appendix~\ref{sec:main_apndx} for a detailed reference). Most notably, the performance improves steadily with generation: when averaging across all instances and repeats, CR increases from $17.5\%$ in generation 1 to $30.0\%$ in generation 10, while the median increases from $11.2\%$ to $26.0\%$. Moreover, all reductions are strictly positive throughout the search, and by generation 10 they range from $5.2\%$ to $72.7\%$. These results provide two main insights. First, \algo{} consistently discovers policies that outperform the numerically optimized base-stock benchmark across a heterogeneous set of inventory instances, indicating that the gains are not driven by only a few favorable settings. Second, most of the improvement occurs early in the evolutionary search: $84.8\%$ of the total increase in the generation-level mean from generations 1 to 10 is realized by generation 5 and $90.6\%$ by generation 6. Thus, the search rapidly identifies effective policy structures in the first few generations and delivers smaller, incremental refinements.

    \begin{figure}[t]
        \centering
        \caption{Impact of lead time (left) and cost parameters (right) on the cost reductions achieved by \algo{}.}
        \includegraphics[width=1\linewidth]{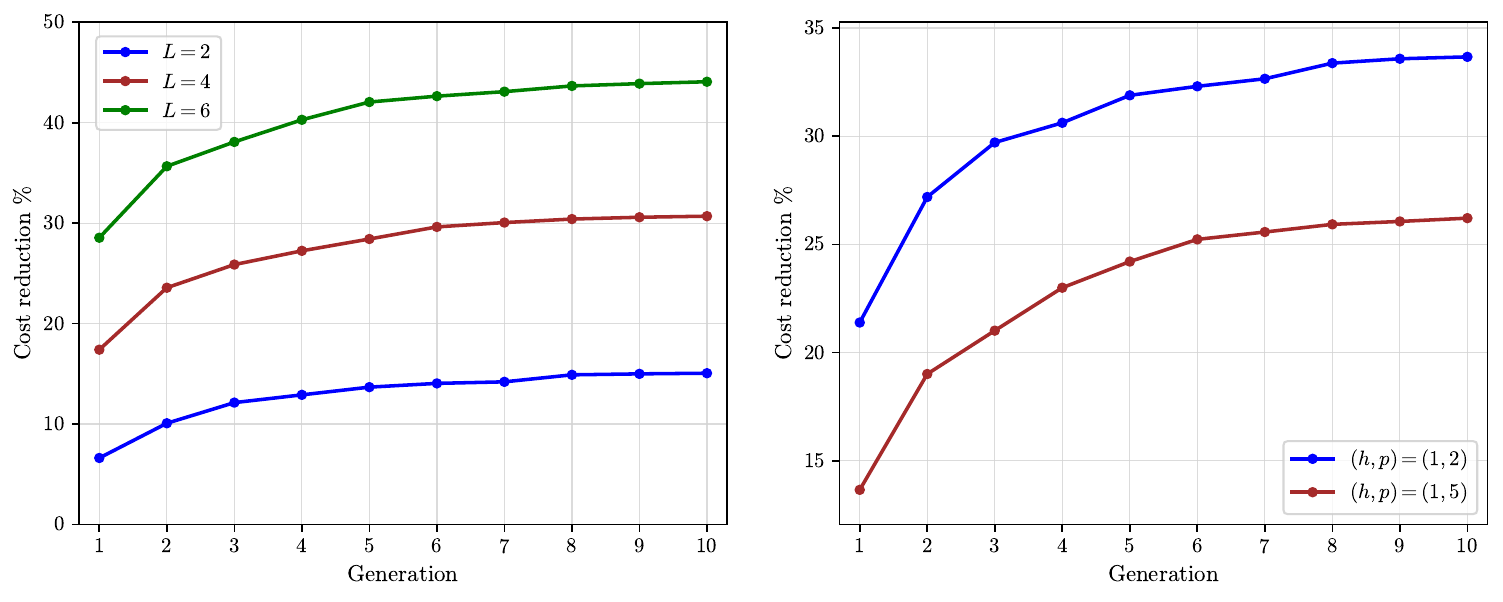}\vspace{-36pt}
        \label{fig:evo_leadtime_cost}
    \end{figure}
    
Next, we examine how the gains from \algo{} vary across problem characteristics. Figure~\ref{fig:evo_leadtime_cost} reports the generation-level mean of cost reduction, with generation on the horizontal axis and cost reduction percentage on the vertical axis. The left panel groups the instances based on lead time: the blue line corresponds to all instances with \(L=2\), the brown line denotes \(L=4\), and the green line denotes \(L=6\). Cost reduction increases with generation for all three lead times, but the magnitude differs substantially: by generation 10, the mean reduction is \(15.1\%\) for \(L=2\), \(30.7\%\) for \(L=4\), and \(44.1\%\) for \(L=6\). The right panel performs the same comparison across cost settings, with the blue line corresponding to all instances with \((h,p)=(1,2)\) and the brown line denotes \((h,p)=(1,5)\). By generation 10, the corresponding mean reductions are \(33.6\%\) and \(26.3\%\), respectively. These patterns indicate that the value of richer policy structures increases markedly with lead time and, to a lesser extent, when holding inventory is relatively more costly than incurring lost sales. One interpretation is that, as lead time grows, the base-stock policy's aggregation of the pipeline into a single inventory-position measure becomes increasingly restrictive, creating greater scope for policies that respond more selectively to the composition of the pipeline. Similarly, the larger gains under \((h,p)=(1,2)\) suggest greater value from refining inventory adjustments when excess inventory is relatively costly. We return to these structural mechanisms in \S\ref{sec:generalization}.

     \begin{figure}[h]
            \centering
            \caption{Density plots of cost reductions achieved by \algo{} at the last generation.}
            \includegraphics[width=0.9\linewidth]{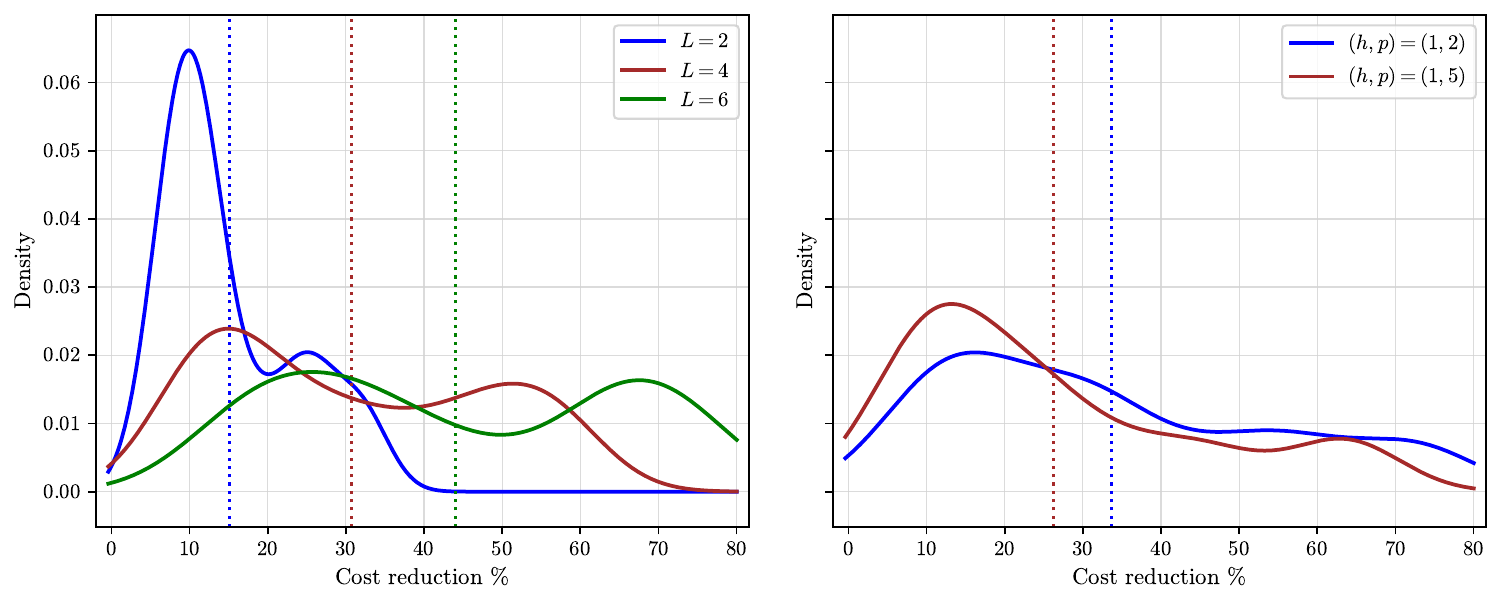}
            \vspace{-6mm}
            \label{fig:density}
    \end{figure}

Figure~\ref{fig:density} complements the generation-level averages in Figure~\ref{fig:evo_leadtime_cost} by showing the fitted density curve and the average value (represented by the dotted lines) of repeat-level cost reductions at the end of the evolutionary search. The left panel echoes the previous finding with a rightward shift as lead time increases: the mean cost reduction rises from (15.1\%) for ($L=2$) to (30.7\%) for ($L=4$) and (44.1\%) for ($L=6$). The broader and partly multimodal densities for (L=4) and (L=6) indicate that the magnitude of the improvement remains heterogeneous across demand environments and repeats; nevertheless, the shift in the overall distributions, rather than only in their means, shows that the larger gains under longer lead times are not driven by a small number of extreme outcomes. This pattern is consistent with the structural interpretation developed in the paper: as the pipeline lengthens, the base-stock policy’s aggregation of all outstanding orders into a single inventory-position measure becomes increasingly restrictive, creating greater value for policies that distinguish among pipeline positions and moderate their replenishment responses. The right panel shows a similar, although less pronounced, effect of the cost parameters: the distribution under ($(h,p)=(1,2)$) is shifted to the right of that under ($(h,p)=(1,5)$), with respective means of (33.7\%) and (26.2\%). Their substantial overlap indicates that demand characteristics and lead time remain important sources of heterogeneity within each cost setting, but the overall shift supports the interpretation that richer policy structures are particularly valuable when holding excess inventory is relatively costly compared with incurring lost sales.

    \begin{figure}[h]
            \centering
            \caption{Impact of demand distributions on the cost reductions achieved by \algo{}.}
            \includegraphics[width=0.95\linewidth]{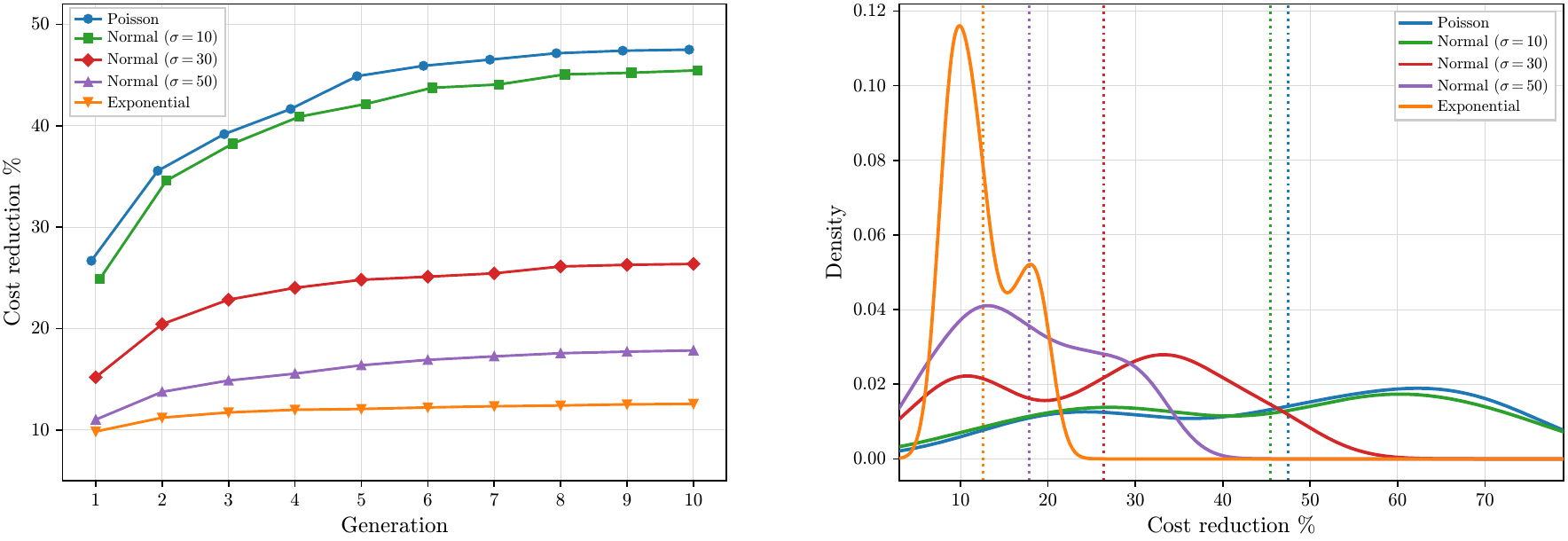}\vspace{-6mm}
            \label{fig:evo_demand}
    \end{figure}

Figure~\ref{fig:evo_demand} examines heterogeneity across demand specifications.  In both panels, the five curves correspond to Poisson demand, Exponential demand,  and Normal demand with standard deviations $\sigma\in\{10,30,50\}$. In the left  panel, mean cost reduction remains positive and increases with generation under all five specifications, but its magnitude varies substantially. By generation  10, the mean reduction reaches approximately $48\%$ under Poisson demand and $45\%$ under Normal demand with $\sigma=10$, compared with $27\%$, $18\%$, and $12\%$ under Normal demand with $\sigma=30$, Normal demand with $\sigma=50$, and Exponential demand, respectively. Within the Normal family, the gains decline  monotonically with demand variability, suggesting that the richer policy structures discovered by \algo{} are particularly valuable when demand is relatively stable. The similar performance under Poisson demand and Normal demand with $\sigma=10$, which have comparable means and standard deviations, further suggests that the discrete-versus-continuous distinction is less important in these low-variability environments. In contrast, the comparatively small gain under Exponential demand is consistent with the greater difficulty of exploiting pipeline information when demand is highly variable and skewed. The right panel shows the fitted densities of $\mathrm{CR}_{k,r}^{(10)}$, with the dotted lines marking the corresponding means. The densities exhibit the same  ordering as the generation-level averages. Thus, the differences in average performance reflect broad shifts in the distributions rather than a small number of unusually successful repeats. The broad and partly multimodal densities also reveal heterogeneity across lead times and cost settings within each demand model.

\subsection{Structural Analysis of Discovered Policies}\label{sec:motif_analysis}

The previous results show that \algo{} can produce policies with substantially lower empirical cost than the optimized base-stock policy across different settings. We now ask what these policies look like. This is important because a discovered policy is most useful when its functional form can be interpreted and related to known replenishment logic. In particular, we aim to answer three questions: First, what recurring design motifs appear in the generated policies? Second, which motifs are most common among top-ranked and later-generation policies? Third, do the discovered policies look like opaque black-box code, or do they recombine interpretable building blocks of inventory policy?

We begin by defining the structural motifs that appear in the discovered policies. We say that a policy contains a given motif if its functional form includes the corresponding structural feature, allowing for algebraically equivalent expressions and differences in variable names across the generated Python code.

\vspace{1mm}
\begin{enumerate}[(a)]
    \item \textbf{Inventory position motif.} The policy depends on the total inventory position 
    $
        \InvOn+\sum_{k=1}^{L}\InvPipe_{t,k}.
    $
    This is the core state statistic used by the classical base-stock policy.

    \vspace{1mm}
    \item \textbf{Pipeline weighting motif.} The policy uses a \textit{linearly weighted} pipeline term 
    $
        \sum_{k=1}^{L}w_k \InvPipe_{t,k}.
    $
    This motif modifies how outstanding orders enter the inventory signal. The base-stock policy gives each pipeline unit the same credit. This aggregation is potentially restrictive, because orders arriving sooner and orders arriving later need not have the same value for avoiding lost sales over the next few periods. Weighted pipeline terms allow the policy to assign different credits to different parts of the pipeline. 

    \vspace{1mm}
    \item \textbf{Nonlinear pipeline composition motif.} The policy computes an inventory signal using a \textit{nonlinear} function \(\varphi(\bm{\InvPipe}_t)\) of the pipeline vector. Examples include \(\varphi(\bm{\InvPipe}_t)=\max_k \InvPipe_{t,k}\), \(\varphi(\bm{\InvPipe}_t)=\min_k \InvPipe_{t,k}\), and \(\varphi(\bm{\InvPipe}_t) = (\sum_{k=1}^{m}\InvPipe_{t,k})/(\sum_{k=1}^{L}\InvPipe_{t,k})\) for some \(m<L\). This motif further generalizes the pipeline weighting motif by introducing non-linearity into the order decision.

    \vspace{1mm}
    \item \textbf{Order-up-to motif.} The order quantity is the \textit{exact non-negative gap} between a fixed target level \(\eta\) and a \textit{state-dependent inventory signal} \(f(\bm{\Inv}_t)\). That is,
    $
        \left(\eta-f(\bm{\Inv}_t)\right)^+.
    $
    This motif (together with the inventory position motif) subsumes the base-stock rule. It introduces a target-seeking negative-feedback structure: when the inventory signal falls below the target, the policy orders enough to close the gap.

    \vspace{1mm}
    \item \textbf{Partial adjustment motif.} The order quantity is \textit{a fraction of} the non-negative gap between a fixed target level and an inventory signal: 
        $\alpha\left(\eta-f(\bm{\Inv}_t)\right)^+,
        \alpha\in(0,1).
    $
    Relative to the order-up-to motif, partial adjustment changes the aggressiveness of the replenishment response. Under base-stock policy, every positive gap is closed immediately. With positive lead times, such one-step correction can be too aggressive: a large order placed today arrives only after a delay, while additional orders may be placed before the effect of the first order is observed. Partial adjustment closes the gap gradually and can reduce overshooting.

    \vspace{1mm}
    \item \textbf{State-dependent order-up-to motif.} This motif generalizes the order-up-to motif by allowing the target to depend on the state:     $
        \left(\eta(\bm{\Inv}_t)-f(\bm{\Inv}_t)\right)^+.
    $
    The target may change with on-hand inventory or with features of the pipeline, rather than remaining fixed across all states.

    \vspace{1mm}
    \item \textbf{State-dependent partial adjustment motif.} This motif combines a state-dependent target with partial adjustment:
    $
        \alpha\left(\eta(\bm{\Inv}_t)-f(\bm{\Inv}_t)\right)^+.
    $
    It allows the policy to change both the target and the strength of the response to current states.

    \vspace{1mm}
    \item \textbf{Constant order motif.} The policy contains a state-independent baseline replenishment component \(b_0\). This component may appear as an additive term, as a lower bound, or as a constant amount ordered in a subset of states. A positive baseline keeps inventory flowing into the pipeline and can prevent the system from waiting too long before restarting replenishment.

    \vspace{1mm}
    \item \textbf{Order clipping motif.} The policy imposes a fixed lower or upper bound on the order quantity, for example,
    $
        \min\{\overline \theta,\max\{\underline \theta,g(\bm{\Inv}_t)\}\}.
    $
    Clipping regularizes extreme order quantities. An upper clip prevents very large corrective orders in depleted states, while a lower clip prevents replenishment from becoming too small in states where a continued inventory flow is desirable.

    \vspace{1mm}
    \item \textbf{Order smoothing motif.} The policy explicitly uses the previous order quantity, which is represented by the last component \(q_{t,L}\) of the current pipeline vector after initialization. Examples include $\min\{g(\bm{\Inv}_t),\InvPipe_{t,L}+\Delta\},$ $\max\{g(\bm{\Inv}_t),\InvPipe_{t,L}-\Delta\},$ and $\beta g(\bm{\Inv}_t)+(1-\beta)\InvPipe_{t,L}.$ This motif avoids abrupt changes in order quantities across adjacent periods while preserving stationarity, because \(\InvPipe_{t,L}\) is part of the current state.

    \vspace{1mm}
    \item \textbf{Thresholding motif.} The policy reacts only when a state-dependent signal crosses a threshold, for example,
    $
        f(\bm{\Inv}_t)\cdot \mathbf{1}\{g(\bm{\Inv}_t)>\tau\}.
    $
    Even without a fixed ordering cost, thresholding can be useful in a lost-sales system with lead times. Small inventory gaps may reflect temporary noise, and reacting to all of them immediately can generate replenishment that arrives after the imbalance has already disappeared.
\end{enumerate}

    Policies generated by \algo{} may differ in algebraic form, parameter names, or code implementation while still sharing the same underlying replenishment logic. We therefore interpret the motifs above as recurring structural building blocks rather than literal code patterns. For example, the LLM may introduce parameters named \texttt{smoothing\_factor}, \texttt{aggressiveness}, or \texttt{order\_scaler}, even though each implements the same partial-adjustment motif. These motifs are not mutually exclusive; most discovered policies combine several of them within a single replenishment rule.

    Viewed through the lens of the base-stock policy, these motifs primarily enrich two aspects of the decision rule. The first is \emph{state representation}. The base-stock policy summarizes the pipeline through the aggregate quantity $\sum_{k=1}^{L}\InvPipe_{t,k}$, thereby assigning the same weight to all outstanding orders regardless of when they arrive. The discovered policies can instead use weighted, nonlinear, or otherwise state-dependent summaries of the pipeline, allowing the ordering decision to distinguish among different pipeline configurations. The second is \emph{inventory adjustment}. A base-stock policy responds to a positive target gap by attempting to close it immediately. The discovered policies can modify this response through partial adjustment, state-dependent targets, clipping, thresholding, smoothing, or baseline replenishment. Thus, the search can change both \emph{how the inventory state is summarized} and \emph{how the resulting signal is translated into an order quantity}.

    More broadly, these motifs can be viewed as structural \emph{basis functions} for constructing policy approximations, analogous to the basis elements used in approximate dynamic programming to obtain tractable approximations of policies or value functions \citep{powell2011approximate}. From this perspective, \algo{} automates an important part of policy approximation: rather than requiring a domain expert to prespecify the relevant structural bases and how they should be combined for a given problem instance, the LLM searches over these building blocks and their combinations, while the numerical optimizer calibrates the associated parameters.

\begin{table}[h]
\centering
\small
\caption{Prevalence of structural motifs in manually labeled generated policies.}\label{tab:motif_prevalence}
\begin{tabular}{p{0.40\linewidth}ccc}
\toprule
\textbf{Motif}   &
\shortstack{\textbf{All}\\\textbf{(\(n=100\))}} &
\shortstack{\textbf{Top-3 Ranked}\\\textbf{(\(n=40\))}} &
\shortstack{\textbf{Generations 8--10}\\\textbf{(\(n=23\))}}\\
\midrule
Inventory position & 77 (77.0\%) & 30 (75.0\%) & 18 (78.3\%) \\
Pipeline weighting & 49 (49.0\%) & 19 (47.5\%) & 12 (52.2\%) \\
Nonlinear pipeline composition & 28 (28.0\%) & 14 (35.0\%) & 9 (39.1\%) \\
Order-up-to & 30 (30.0\%) & 11 (27.5\%) & 6 (26.1\%) \\
Partial adjustment & 27 (27.0\%) & 8 (20.0\%) & 6 (26.1\%) \\
State-dependent order-up-to & 17 (17.0\%) & 8 (20.0\%) & 4 (17.4\%) \\
State-dependent partial adjustment & 33 (33.0\%) & 15 (37.5\%) & 11 (47.8\%) \\
Constant order & 13 (13.0\%) & 5 (12.5\%) & 5 (21.7\%) \\
Order clipping & 37 (37.0\%) & 15 (37.5\%) & 13 (56.5\%) \\
Order smoothing & 7 (7.0\%) & 4 (10.0\%) & 3 (13.0\%) \\
Thresholding & 13 (13.0\%) & 3 (7.5\%) & 2 (8.7\%) \\
\bottomrule
\end{tabular}

\vspace{0.5em}
\begin{minipage}{0.94\linewidth}
\footnotesize
\emph{Notes: Entries report the number of manually labeled policies containing each motif, with the corresponding percentage in parentheses. Because a policy may contain multiple motifs, the percentages need not sum to 100.}
\end{minipage}
\end{table}\vspace{-10pt}

    Table~\ref{tab:motif_prevalence} summarizes a manual coding exercise designed to identify which structural motifs recur in the policies generated by \algo{}. We first randomly sample 100 generated policies. For each sampled policy, we manually inspect its Python code to understand the proposed functional form, simplify this replenishment logic when necessary, and record a binary indicator for whether each motif defined above is present. A motif is counted only if it remains active after inspecting its optimized parameter; for example, a pipeline-weighting term with an optimized coefficient of zero is treated as absent. The \emph{All} column reports motif frequencies across these 100 sampled policies. We then examine two subsets of the same sample. The \emph{Top-3 Ranked} column includes sampled policies that rank among the three lowest-cost policies within their generation, while the \emph{Generations 8--10} column includes sampled policies generated in the final three generations. Comparing these columns provides a descriptive view of which motifs are common overall and which appear more frequently among better-performing or later-generation policies. Because the two subsets are defined from the same sample, they may overlap. Four patterns stand out:
    \begin{itemize}
        \item The discovered policies often retain inventory position but modify the associated ordering logic. The inventory-position motif appears in \(77.0\%\) of the full sample and remains similarly prevalent among top-ranked policies (\(75.0\%\)) and policies from generations 8--10 (\(78.3\%\)). In contrast, the standard order-up-to motif appears in only \(30.0\%\) of the full sample. Thus, \algo{} frequently preserves inventory position as a useful state summary while changing how that information is translated into an order quantity.
    
        \item Richer representations of pipeline inventory are common. Pipeline weighting appears in \(49.0\%\) of the full sample, while nonlinear pipeline composition appears in \(28.0\%\). Nonlinear pipeline composition is more prevalent among top-ranked policies (\(35.0\%\)) and policies from generations 8--10 (\(39.1\%\)). These patterns suggest that many discovered policies distinguish among different parts of the pipeline rather than assigning equal importance to all outstanding orders. This provides one possible explanation for the lead-time results in Figure~\ref{fig:evo_leadtime_cost}, where the gains from \algo{} increase substantially with lead time.
    
        \item The discovered policies frequently moderate the replenishment response. Order clipping appears in \(37.0\%\) of the full sample and in \(56.5\%\) of policies from generations 8--10. Similarly, state-dependent partial adjustment appears in \(33.0\%\) of the full sample, \(37.5\%\) of top-ranked policies, and \(47.8\%\) of policies from generations 8--10. These results indicate that \algo{} often retains target-seeking behavior while regulating how aggressively the policy responds to an inventory shortfall. By comparison, order smoothing and thresholding are less common, appearing in \(7.0\%\) and \(13.0\%\) of the full sample, respectively, suggesting that they serve as more specialized refinements.
    
        \item The motifs typically appear in combination rather than in isolation. A sampled policy contains \(3.31\) motifs on average, the median policy contains three motifs, and only 3 of the 100 sampled policies contain a single motif. Thus, the policies discovered by \algo{} are best viewed as combinations of a small set of interpretable inventory-control building blocks rather than as arbitrary black-box rules or minor variations of a single benchmark policy. Overall, three recurring ingredients emerge: informative state summaries, richer representations of pipeline inventory, and moderated replenishment responses.
    \end{itemize}

\subsection{Representative Evolutionary Trajectory} \label{represent_trajectory}

\begin{figure}[t]
        \centering
        \caption{Evolution of the best retained policy for a representative instance and repeat.\vspace{2pt}}
        \includegraphics[width=\linewidth]
        {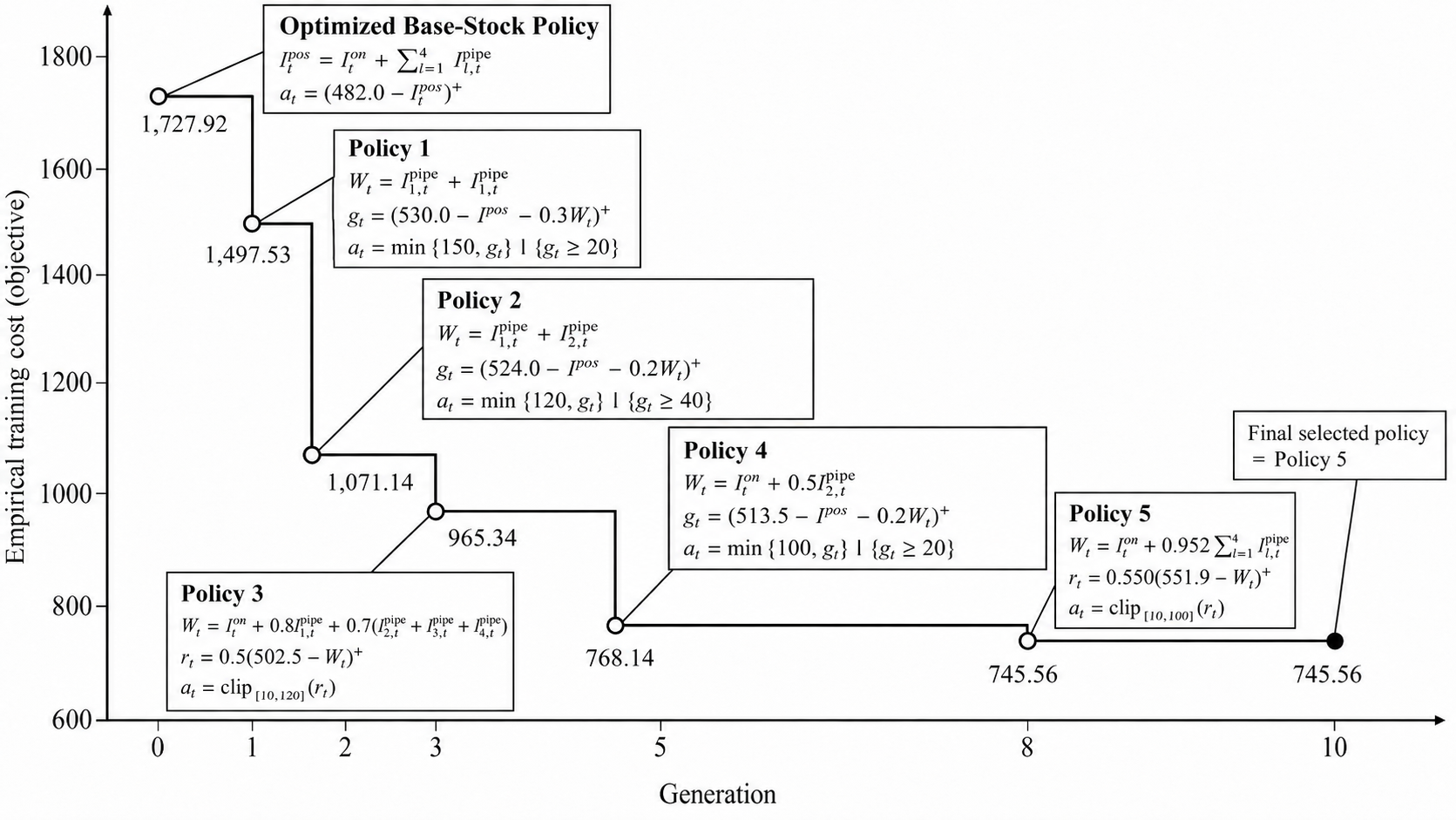}\vspace{-30pt}
        \label{fig:representative_policy_trajectory}
    \end{figure}

The previous section summarizes which structural motifs recur across policies generated by \algo{}, but it does not show how the search moves from one policy to another within a particular run. Figure~\ref{fig:representative_policy_trajectory} therefore traces one representative repeat for the Poisson-demand instance with lead time \(L=4\) and cost parameters \((h,p)=(1,2)\). The horizontal axis denotes the generation, and the vertical axis reports the empirical training cost of the best policy retained up to that generation. Generation 0 corresponds to the optimized base-stock policy used to initialize the search. Whenever \algo{} discovers a policy with lower empirical training cost than the current incumbent, the figure reports the new policy's simplified functional form in an accompanying box. In this run, new policies are discovered in generations 1, 2, 3, 5, and 8, corresponding to Policies 1--5, respectively. The empirical cost decreases from \(1{,}727.92\) for the optimized base-stock benchmark to \(1{,}497.53\), \(1{,}071.14\), \(965.34\), \(768.14\), and ultimately \(745.56\). No better policy is found after generation 8, so Policy~5 remains the best candidate through generation 10 and is returned as the final selected policy.  

    To illustrate how the policy annotations in Figure~\ref{fig:representative_policy_trajectory} should be interpreted, consider Policy~1, the first policy to improve on the optimized base-stock benchmark. Since \(L=4\), denote the inventory position and the quantity arriving in the first two pipeline positions as
    \(
        I_t^{\mathrm{pos}}:=\InvOn_t+\sum_{\ell=1}^{4}\InvPipe_{t,\ell}, W_t:=\InvPipe_{t,1}+\InvPipe_{t,2}.
    \)
    Policy~1 computes the adjusted target gap $g_t:=\left(530.0-I_t^{\mathrm{pos}}-0.3W_t\right)^+$, and places the order $a_t:=\min\{150,g_t\}\mathbf{1}\{g_t\geq20\}$. Relative to the base-stock policy, this rule changes both the representation of the inventory state and the response to the resulting target gap. Because \(I_t^{\mathrm{pos}}\) already assigns unit weight to every pipeline component, subtracting an additional \(0.3W_t\) gives the first two pipeline positions a total weight of \(1.3\), while the remaining two retain weight \(1\). Equivalently,
    \[
    g_t=\left(530.0-\InvOn_t-1.3\InvPipe_{t,1}-1.3\InvPipe_{t,2}-\InvPipe_{t,3}-\InvPipe_{t,4}\right)^+.
    \]
    Thus, Policy~1 distinguishes between near- and later-arriving pipeline inventory rather than treating all outstanding orders identically. It also modifies the base-stock policy's one-step adjustment through a threshold and an order cap:
    \[
    a_t=
    \begin{cases}
    0, & \text{if } g_t<20,\\[2pt]
    g_t, & \text{if } 20\leq g_t\leq150,\\[2pt]
    150, & \text{if } g_t>150.
    \end{cases}
    \]
    The threshold prevents small target gaps from triggering an order, while the upper bound limits excessively large corrective orders. Policy~1 therefore combines the inventory-position, pipeline-weighting, order-up-to, thresholding, and order-clipping motifs, and reduces the empirical training cost from \(1{,}727.92\) to \(1{,}497.53\).

    Policy~2 retains essentially the same structural template as Policy~1 but uses different optimized parameter values, reducing the empirical training cost further to \(1{,}071.14\). This illustrates that two generated policies can share the same motif combination yet attain different performance because the numerical optimizer identifies different parameterizations under its finite computational budget. The subsequent incumbents introduce more substantial structural changes. Policy~3 replaces the standard inventory-position signal with a timing-sensitive discounted representation of the pipeline and combines this signal with partial adjustment and two-sided order clipping. Policy~4 returns to inventory position but augments it with a timing-weighted near-arrival signal and an upper order cap. Finally, Policy~5 uses a uniformly discounted pipeline signal together with partial adjustment and two-sided clipping. Its empirical training cost is \(745.56\), corresponding to a \(56.9\%\) reduction relative to the optimized base-stock benchmark. Policy~5 becomes the incumbent in generation 8 and remains the best retained policy through generation 10.

    This trajectory illustrates how the structural motifs identified in Table~\ref{tab:motif_prevalence} emerge within an actual search run. The search repeatedly preserves the basic target-seeking logic of inventory control while changing two key components: how the pipeline state is summarized and how aggressively the resulting target gap is converted into an order. Importantly, the evolutionary path does not simply add complexity from one generation to the next. Instead, it moves among different combinations of pipeline weighting, partial adjustment, thresholding, and clipping before arriving at a relatively parsimonious final rule.

\section{Cross-Environment Generalization}\label{sec:generalization}

    This section studies whether the policy classes discovered by \algo{} generalize beyond the problem instances on which they were generated. Importantly, the object transferred across instances is the policy class, not a particular policy with fixed parameter values. Specifically, for a discovered class \(\Pi_{\Theta}:=\{\pi_{\bm{\theta}}:\bm{\theta}\in\Theta\}\), we retain its functional form \(\pi_{\bm{\theta}}(\cdot)\) and feasible parameter set \(\Theta\), and re-optimize \(\bm{\theta}\) using training data from a new target instance that differs in its demand distribution, cost parameters, and/or lead time from the instance on which the class was discovered. We then evaluate the resulting optimized policy against the optimized base-stock policy for that same target instance. This cross-environment test therefore measures \emph{structural generalization}: whether a functional form discovered in one environment continues to contain high-performing policies after its parameters are recalibrated for another environment. In this way, we assess whether the structural motifs identified in \S\ref{sec:motif_analysis} combine into reusable policy classes rather than instance-specific policies.\looseness=-1

    We use three reference instances to generate the policy classes studied in this section. All three share the same lead time and cost ratio, \(L=6\) and \(R:=p/h=2\), and differ only in the demand distribution. The first has Poisson demand with mean \(\mu=100\) and standard deviation \(\sigma=10\), corresponding to a low-variability environment with coefficient of variation \(\sigma/\mu=10\%\). The second has Normal demand with \(\mu=100\) and \(\sigma=30\), representing moderate variability with coefficient of variation \(30\%\). The third has Exponential demand with \(\mu=100\) and \(\sigma=100\), representing high variability with coefficient of variation \(100\%\). In \S\ref{sec:llm-policies}, we describe the three policy classes discovered from these reference instances and relate their structures to the motifs identified in \S\ref{sec:motif_analysis}. In \S\ref{sec:test-instance}, we define the target instances and performance metrics. In \S\ref{sec:results}, we report the generalization results.

\begin{table}[h]
\centering
\small
\caption{Structural motifs in the three discovered policy classes.}
\label{tab:discovered_policy_motifs}

\resizebox{\textwidth}{!}{%
\begin{tabular}{p{0.2\linewidth}p{0.2\linewidth}p{0.65\linewidth}}
\toprule
\textbf{Reference demand} & \textbf{Policy class} & \textbf{Main motifs} \\
\midrule
Poisson & Capped & Inventory position, order-up-to, order clipping \\
Exponential & Discounted-pipeline & Pipeline weighting, partial adjustment, thresholding \\
Normal & Hybrid & Pipeline weighting, partial adjustment, constant order, order clipping \\
\bottomrule
\end{tabular}%
}
\end{table}\vspace{-20pt}

\subsection{Discovered Inventory Policy Classes}\label{sec:llm-policies}

On the three reference instances, \algo{} discovers three policy classes that we use in the generalization analysis: the capped, discounted-pipeline, and hybrid classes. We next describe their functional forms and relate them to the structural motifs identified in \S\ref{sec:motif_analysis}. For each \(q\in\{\mathrm C,\mathrm D,\mathrm H\}\), let \(\Pi_{\Theta_q}^{q}:=\{\pi_{\bm\theta_q}^{q}:\bm\theta_q\in\Theta_q\}\) denote the corresponding discovered class, where \(\bm\theta_q\) is its parameter vector and \(\Theta_q\) is the feasible parameter space returned by the LLM. Table~\ref{tab:discovered_policy_motifs} summarizes the reference instance on which each class was discovered and its main structural motifs. For any state \(\Inv\), let \(I^{\mathrm{pos}}:=\InvOn+\sum_{k=1}^{L}\InvPipe_k\) denote the inventory position, following \S\ref{sec:model}. We suppress the common upper bound \(\bar a\) on feasible order quantities because it is chosen sufficiently large to be nonbinding.

\textbf{Capped policy class.}
The class discovered on the Poisson reference instance preserves the inventory-position representation and target-gap logic of the base-stock policy, but caps the resulting order quantity. Let \(\bm\theta_{\mathrm C}:=(\theta_{\mathrm C},\kappa_{\mathrm C})\in\Theta_{\mathrm C}\), where \(\theta_{\mathrm C}\) is the target inventory position and \(\kappa_{\mathrm C}\) is the order cap. For any state \(\Inv\), the policy is \(\pi_{\bm\theta_{\mathrm C}}^{\mathrm C}(\Inv):=\min\big\{\kappa_{\mathrm C},\big(\theta_{\mathrm C}-I^{\mathrm{pos}}\big)^+\big\}\). Thus, relative to the base-stock class, the state representation is unchanged; the structural modification is the order-clipping motif, which prevents a large target gap from generating an equally large corrective order. When \(\kappa_{\mathrm C}\) is nonbinding, the policy reduces to the standard base-stock rule.

\textbf{Discounted-pipeline policy class.}
The class discovered on the Exponential reference instance modifies both how pipeline inventory is represented and how an inventory shortfall is translated into an order. Let \(\bm\theta_{\mathrm D}:=(\theta_{\mathrm D},\omega_{\mathrm D},\rho_{\mathrm D},\tau_{\mathrm D})\in\Theta_{\mathrm D}\), and define \(x_{\mathrm D}(\Inv):=\InvOn+\omega_{\mathrm D}\sum_{k=1}^{L}\InvPipe_k\) and \(g_{\mathrm D}(\Inv):=\theta_{\mathrm D}-x_{\mathrm D}(\Inv)\). For any state \(\Inv\), the policy is \(\pi_{\bm\theta_{\mathrm D}}^{\mathrm D}(\Inv):=\rho_{\mathrm D}\big(g_{\mathrm D}(\Inv)\big)^+\mathbf{1}\{g_{\mathrm D}(\Inv)>\tau_{\mathrm D}\}\). Here, \(\omega_{\mathrm D}\) determines the weight assigned to pipeline inventory, \(\rho_{\mathrm D}\) determines the fraction of the target gap that is replenished, and \(\tau_{\mathrm D}\) creates an inaction region for small shortfalls. Thus, relative to the base-stock class, the discounted-pipeline class modifies both the state representation and the replenishment response through the pipeline-weighting, partial-adjustment, and thresholding motifs.

\textbf{Hybrid policy class.}
The class discovered on the Normal reference instance combines position-specific pipeline weighting with a moderated replenishment response and a state-independent baseline component. Let \(\bm\theta_{\mathrm H}:=(\theta_{\mathrm H},\bm\omega,\alpha,\bm\beta,\rho_{\mathrm H},\delta_{\mathrm H},\underline a_{\mathrm H})\in\Theta_{\mathrm H}\), where \(\bm\omega=(\omega_1,\ldots,\omega_L)\) and \(\bm\beta=(\beta_1,\ldots,\beta_L)\). Define the effective weight on pipeline position \(k\) as \(v_k:=\omega_k-\alpha\beta_k\) and the corresponding inventory signal as \(x_{\mathrm H}(\Inv):=\InvOn+\sum_{k=1}^{L}v_k\InvPipe_k\). For any state \(\Inv\), the policy is \(\pi_{\bm\theta_{\mathrm H}}^{\mathrm H}(\Inv):=\max\big\{\underline a_{\mathrm H},\rho_{\mathrm H}\big(\theta_{\mathrm H}-x_{\mathrm H}(\Inv)\big)+(1-\rho_{\mathrm H})\delta_{\mathrm H}\big\}\). The effective weights \(v_k\) allow different pipeline positions to receive different credits, \(\rho_{\mathrm H}\) moderates the response to the target gap, \((1-\rho_{\mathrm H})\delta_{\mathrm H}\) provides a state-independent replenishment component, and \(\underline a_{\mathrm H}\) imposes a lower order clip. Thus, the hybrid class combines the pipeline-weighting, partial-adjustment, constant-order, and order-clipping motifs within a single replenishment rule.

Taken together, the three classes illustrate different ways in which \algo{} recombines the structural building blocks identified in \S\ref{sec:motif_analysis}. The capped class preserves the base-stock state representation and modifies only the replenishment response, whereas the discounted-pipeline and hybrid classes modify both how pipeline inventory is represented and how the resulting inventory signal is translated into an order. These functional forms, rather than their parameter values on the reference instances, are the objects transferred in the experiments below: for each target environment, we retain the class structure and reoptimize its parameters before evaluating performance.

\subsection{Target Instances and Performance Metrics}\label{sec:test-instance}

We evaluate whether the three policy classes discovered on the reference instances remain effective across a broad collection of target environments. Each target instance \(k\) is characterized by a demand family \(\mathcal F_k\), mean \(\mu_k\), standard deviation \(\sigma_k\), lead time \(L_k\), and lost-sales-penalty-to-holding-cost ratio \(R_k:=p_k/h_k\). We consider 13 demand families spanning bounded continuous distributions (Beta, continuous Uniform, and Triangular), bounded discrete distributions (Binomial and discrete Uniform), light- to moderate-tailed unbounded distributions (Gamma, Geometric, Negative Binomial, Normal, and Weibull), and heavy-tailed or highly skewed distributions (Lognormal, Pareto, and zero-inflated Negative Binomial). Varying the distribution parameters produces 629 distinct demand specifications \((\mathcal F,\mu,\sigma)\); the complete construction is reported in Appendix~\ref{app:demand_generation}. Combining these specifications with \(L\in\{2,4,6,8\}\) and \(R\in\{2,4,6,10\}\) yields \(629\times4\times4=10{,}064\) target instances.

For each target instance, we transfer the three policy classes described in \S\ref{sec:llm-policies} while reoptimizing their parameters for the new environment. Specifically, for each policy class \(q\in\mathcal \{\mathrm C,\mathrm D,\mathrm H\}\), the functional form and feasible parameter space \(\Pi_{\Theta_q}^{q}:=\{\pi_{\bm\theta_q}^{q}:\bm\theta_q\in\Theta_q\}\) are held fixed, and the numerical optimization procedure in \S\ref{sec:parameter-optimization} is applied to obtain an instance-specific parameter vector \(\widehat{\bm\theta}_{k,q}\) for target instance \(k\). We similarly reoptimize the base-stock parameter to obtain \(\widehat\theta_{k,\mathrm B}\). Thus, the four policies compared on instance \(k\) are \(\pi_{\widehat{\bm\theta}_{k,\mathrm C}}^{\mathrm C}\), \(\pi_{\widehat{\bm\theta}_{k,\mathrm D}}^{\mathrm D}\), \(\pi_{\widehat{\bm\theta}_{k,\mathrm H}}^{\mathrm H}\), and \(\pi_{\widehat\theta_{k,\mathrm B}}^{\mathrm{BS}}\). Thus, our experiments evaluate the structural generalization: whether a class discovered in one environment continues to contain effective policies when the demand distribution, lead time, or cost ratio changes.

Following \S\ref{sec:main}, we use percentage cost reduction relative to the optimized base-stock policy as the primary performance metric. For each target instance \(k\) and policy class \(q\), let \(\mathrm{CR}_{k}^{q}\) denote the percentage reduction in empirical cost achieved by the re-optimized policy from class \(q\) relative to the optimized base-stock policy on the same instance. Thus, \(\mathrm{CR}_{k}^{q}>0\) indicates that class \(q\) yields a lower-cost policy than the base-stock benchmark, with larger values indicating greater improvement. To compare policy classes across the \(K=10{,}064\) target instances, we examine both the mean cost reduction and the distribution of \(\mathrm{CR}_{k}^{q}\) across instances. As a complementary measure, we rank the four optimized policies---capped, discounted-pipeline, hybrid, and base-stock---within each target instance according to empirical cost. Let \(\mathrm{PR}_{k}^{q}\in\{1,2,3,4\}\) denote the policy rank, where rank \(1\) corresponds to the lowest empirical cost and rank \(4\) to the highest. We similarly report the mean policy rank and its distribution across target instances. Thus, lower values indicate better relative performance, with a rank close to \(1\) meaning that the policy is typically the best-performing of the four alternatives.\looseness=-1

\subsection{Results}\label{sec:results}

    We begin with the aggregate performance across all \(10{,}064\) target instances. The capped, discounted-pipeline, and hybrid classes achieve mean cost reductions of \(22.60\%\), \(21.75\%\), and \(22.10\%\), respectively, relative to the optimized base-stock benchmark. Their corresponding mean policy ranks are \(1.93\), \(2.17\), and \(2.01\), compared with \(3.89\) for the base-stock policy. Thus, all three discovered classes generalize strongly across environments that differ from the reference instances: after reoptimizing their parameters for each target instance, they reduce cost by more than \(21\%\) on average and typically rank well ahead of the base-stock benchmark. Among the three discovered classes, the capped class performs best on average, with the highest mean cost reduction and lowest mean policy rank, followed closely by the hybrid and discounted-pipeline classes. More broadly, the key result is not that a single discovered class dominates across all environments, but that each transferred functional form continues to contain high-performing policies across a wide range of demand distributions, lead times, and cost ratios.

\begin{figure}[h]
    \centering
    \caption{Distribution of cost reduction percentages across lead times and policy classes.}
    \includegraphics[width=\linewidth]{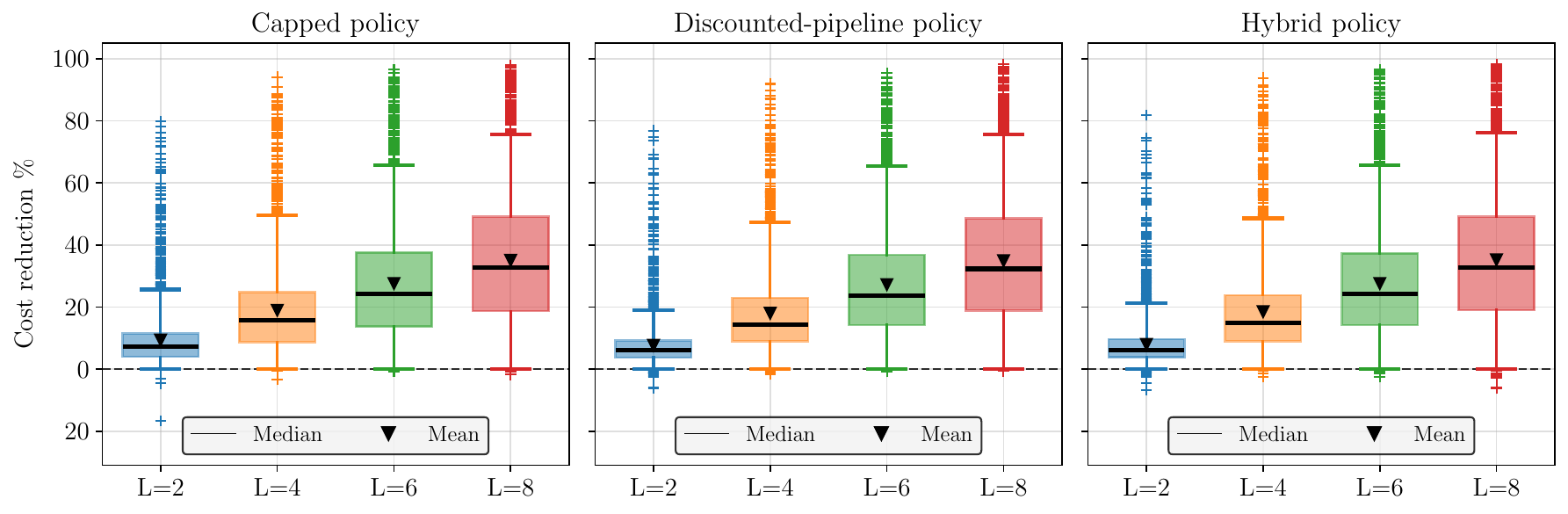}
    \label{fig:leadtime_comparison}
    \vspace{-36pt}
\end{figure}

    \textbf{Impact of lead time variation.} Figure~\ref{fig:leadtime_comparison} shows how the distribution of cost reduction changes with lead time for the capped, discounted-pipeline, and hybrid policy classes. The left, middle, and right panels correspond to the three classes, respectively. In each panel, the vertical axis reports cost reduction relative to the optimized base-stock policy, and the boxplots correspond to \(L\in\{2,4,6,8\}\). The solid line denotes the median and the triangle denotes the mean. Across all three classes, the distributions shift markedly upward as lead time increases. For the capped class, mean cost reduction rises from \(9.20\%\) at \(L=2\) to \(18.86\%\), \(27.43\%\), and \(34.90\%\) at \(L=4,6,8\), respectively. The corresponding values are \(7.47\%\), \(17.84\%\), \(27.00\%\), and \(34.70\%\) for the discounted-pipeline class, and \(7.80\%\), \(18.32\%\), \(27.32\%\), and \(34.94\%\) for the hybrid class. Moreover, Table~\ref{tab:leadtime_mpr} provides a complementary comparison based on relative ranking. All three discovered policy classes rank substantially better than the base-stock benchmark, whose mean policy rank remains close to \(4\) for every lead time. Among the discovered classes, the capped class ranks best at \(L=2\) and \(L=4\), while the hybrid class has the lowest mean policy rank at \(L=6\) and \(L=8\). Thus, although all three classes perform similarly at longer lead times, the hybrid class gains a slight relative advantage as the pipeline lengthens.\looseness=-1

    \begin{table}[h]
        \centering
        \caption{Mean policy rank across lead times.}
        \label{tab:leadtime_mpr}
        \begin{tabular*}{\linewidth}{@{\extracolsep{\fill}}c c @{\hspace{0.5cm}} c @{\hspace{0.5cm}} c @{\hspace{0.5cm}} c @{}}
            \toprule
            $L$ & Capped & Discounted-pipeline & Hybrid & Base-stock \\
            \midrule
            2 & \textbf{1.78} & 2.24 & 2.12 & 3.86 \\
            4 & \textbf{1.87} & 2.24 & 1.99 & 3.89 \\
            6 & 1.99 & 2.13 & \textbf{1.98} & 3.90 \\
            8 & 2.06 & 2.07 & \textbf{1.95} & 3.91 \\
            \bottomrule
        \end{tabular*}
        \vspace{-20pt}
    \end{table}

    Together, Figure~\ref{fig:leadtime_comparison} and Table~\ref{tab:leadtime_mpr} show that the transferred policy classes substantially outperform the optimized base-stock benchmark across the full lead-time range, with the advantage becoming especially pronounced for long lead times. The upward shift in the full cost-reduction distributions indicates that this pattern is not driven by only a few favorable target instances. One interpretation is that, as lead time increases, the base-stock policy's aggregation of all pipeline inventory into a single inventory-position measure becomes increasingly restrictive, creating greater value for policy classes that represent the pipeline more flexibly and moderate the replenishment response.

\begin{table}[h]
    \centering
    \caption{Mean cost reduction and mean policy rank across lost-sales-cost-to-holding-cost ratios \(R=p/h\).}
    \label{tab:cost_generalization}
    \resizebox{\linewidth}{!}{%
    \begin{tabular}{c@{\hspace{1cm}} cc c@{\hspace{1cm}} cc c@{\hspace{1cm}} cc c@{\hspace{1cm}} cc}
        \toprule
        & \multicolumn{2}{c}{Capped}
        & &
        \multicolumn{2}{c}{Discounted-pipeline}
        & &
        \multicolumn{2}{c}{Hybrid}
        & &
        \multicolumn{2}{c}{Base-stock} \\
        \cmidrule(lr){2-3}
        \cmidrule(lr){5-6}
        \cmidrule(lr){8-9}
        \cmidrule(lr){11-12}
        \(R\)
        & \shortstack{Cost\\reduction }
        & \shortstack{Policy\\rank}
        & 
        & \shortstack{Cost\\reduction }
        & \shortstack{Policy\\rank}
        &
        & \shortstack{Cost\\reduction }
        & \shortstack{Policy\\rank}
        &
        & \shortstack{Cost\\reduction }
        & \shortstack{Policy\\rank} \\
        \midrule
        2  & 26.93 & 2.22 & & 26.80 & 2.04 & & \textbf{26.95} & \textbf{1.90} & & -- & 3.84 \\
        4  & \textbf{23.35} & \textbf{1.98} & & 22.84 & 2.14 & & 23.11 & \textbf{1.98} & & -- & 3.91 \\
        6  & \textbf{21.30} & \textbf{1.82} & & 20.29 & 2.24 & & 20.69 & 2.05 & & -- & 3.89 \\
        10 & \textbf{18.81} & \textbf{1.68} & & 17.08 & 2.27 & & 17.63 & 2.12 & & -- & 3.92 \\
        \bottomrule
    \end{tabular}%
    }
\end{table}

\textbf{Impact of cost ratio.} Table~\ref{tab:cost_generalization} reports mean cost reduction and mean policy rank for each policy class across the lost-sales-cost-to-holding-cost ratios \(R\in\{2,4,6,10\}\). For all three discovered classes, mean cost reduction decreases monotonically as \(R\) increases. The capped class declines from \(26.93\%\) at \(R=2\) to \(18.81\%\) at \(R=10\); the discounted-pipeline class declines from \(26.80\%\) to \(17.08\%\); and the hybrid class declines from \(26.95\%\) to \(17.63\%\). Thus, the advantage over the optimized base-stock benchmark is largest when lost sales are relatively inexpensive compared with holding inventory and narrows as \(R\) increases. The mean policy rank results tell the same broad story: the base-stock benchmark remains close to rank \(4\), with values between \(3.84\) and \(3.92\), while the discovered classes remain near rank \(2\). Among the discovered classes, the hybrid class performs best at \(R=2\), the capped and hybrid classes tie at \(R=4\), and the capped class has the lowest mean policy rank at \(R=6\) and \(R=10\).\looseness=-1

The key insight is that the transferred policy classes remain substantially better than the optimized base-stock benchmark across the full range of cost ratios, although the size of the improvement declines as lost sales become more expensive. One interpretation is that a higher \(R\) makes the base-stock policy's aggressive protection against stockouts more competitive, leaving less room for improvement. At the same time, the shift in relative performance across discovered classes is consistent with their structural differences: the hybrid class performs particularly well when holding inventory is relatively costly, whereas the capped class becomes relatively stronger as the lost-sales penalty increases. These patterns suggest that different combinations of the motifs identified in \S\ref{sec:motif_analysis} are better suited to different cost regimes.

\begin{figure}[t]
        \centering
        \caption{Distribution of cost reduction percentages across demand families and policy classes.}
        \includegraphics[width=\linewidth]{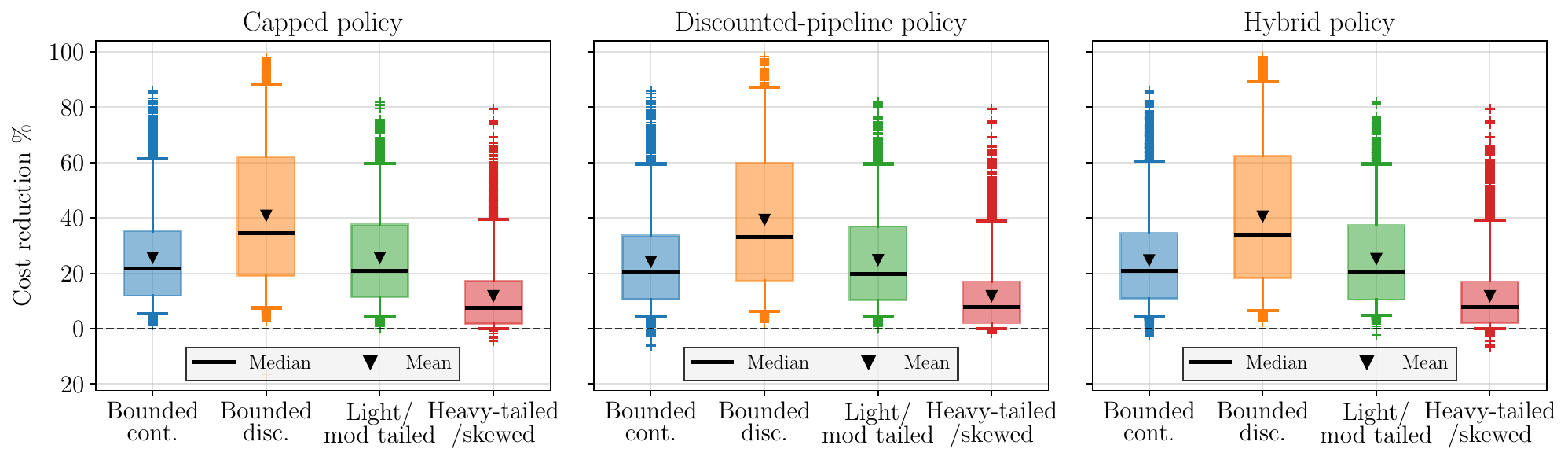}
        \label{fig:demand_comparison}
        \vspace{-32pt}
    \end{figure}

\textbf{Impact of demand distribution.} Figure~\ref{fig:demand_comparison} shows how cost reduction varies across four broad demand groups: bounded continuous, bounded discrete, light- to moderate-tailed, and heavy-tailed or highly skewed distributions. The left, middle, and right panels correspond to the capped, discounted-pipeline, and hybrid policy classes, respectively. In each panel, the vertical axis reports percentage cost reduction relative to the optimized base-stock policy, while the solid line and triangle denote the median and mean. All three classes exhibit a similar pattern across demand environments. Cost reductions are largest under bounded discrete demand, remain substantial under bounded continuous and light- to moderate-tailed demand, and are markedly smaller under heavy-tailed or highly skewed demand. The latter group also contains observations near or below zero for all three classes, indicating that the advantage over base-stock is less consistent in these environments. Overall, the dominant variation is across demand environments rather than across the three transferred classes. These results suggest that the discovered structures transfer particularly well to bounded and moderately behaved demand settings, whereas heavy tails and strong skewness make their performance gains less uniform across target instances.

    \begin{figure}[h]
    \centering
    \caption{Distribution of policy ranks across demand-variability and lead-time regimes.}
    \includegraphics[width=\linewidth]{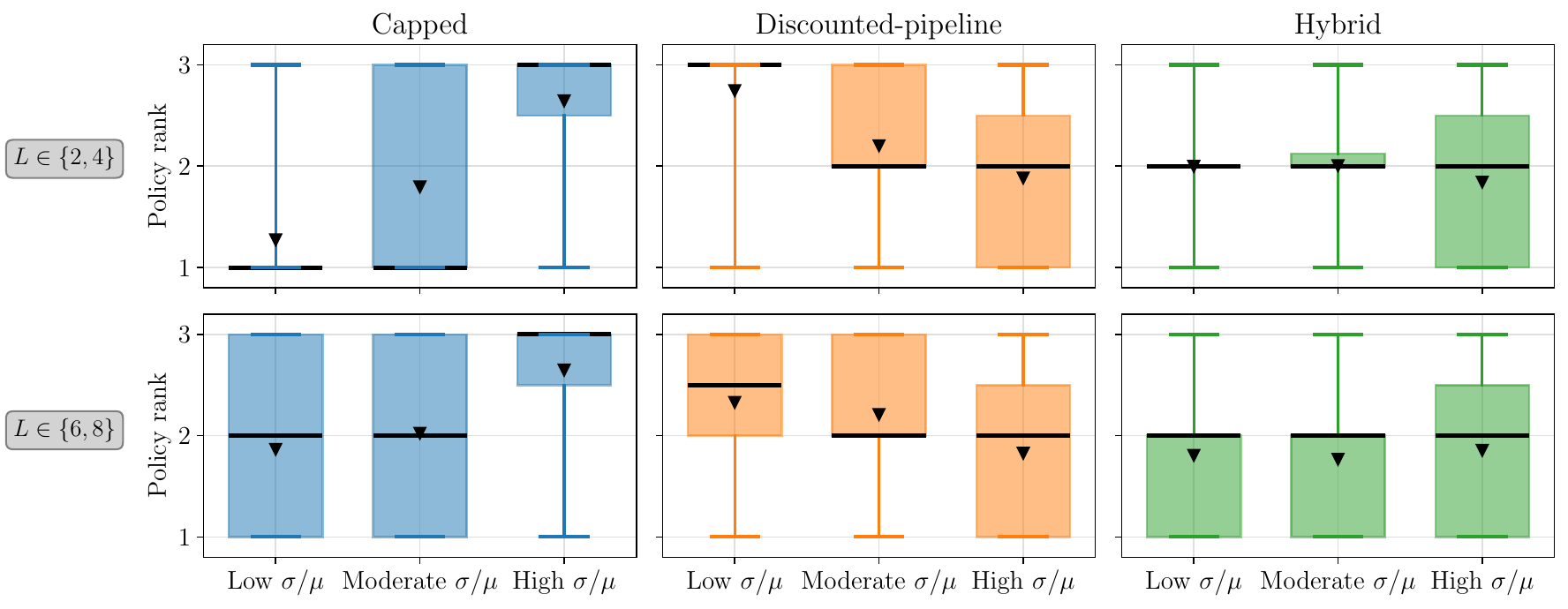}
    \label{fig:rank_cv_lead}\vspace{-38pt}
\end{figure}
Figure~\ref{fig:rank_cv_lead} provides a more direct comparison among the three discovered classes across demand variability and lead time. The columns correspond to the capped, discounted-pipeline, and hybrid classes, while the top and bottom rows correspond to short lead times \(L\in\{2,4\}\) and long lead times \(L\in\{6,8\}\), respectively. Within each panel, the horizontal axis distinguishes low, moderate, and high coefficients of variation \(\sigma/\mu\), and the vertical axis reports policy rank among the three discovered classes, with lower values indicating better relative performance. The capped class performs particularly well under low variability, especially at short lead times, where its median rank is \(1\), but its relative performance deteriorates as variability increases. In contrast, the discounted-pipeline and hybrid classes become more competitive under high variability and substantially outperform the capped class in relative rank. The hybrid class also becomes relatively stronger as lead time increases, particularly under moderate variability.

Taken together, the two figures show that no single transferred class is uniformly best. When demand is relatively stable, retaining the standard inventory-position representation while limiting large corrective orders, as in the capped class, is often sufficient. As variability increases, richer treatment of pipeline inventory becomes more valuable, favoring the discounted-pipeline and hybrid classes. Longer lead times further increase the value of distinguishing pipeline information and moderating the replenishment response. Thus, the relative performance of the discovered classes varies systematically with the demand environment and pipeline length, providing evidence that different combinations of structural motifs are useful in different operating regimes.

\section{Value of Optimization}\label{sec:external}

The previous sections use \algo{} with an external numerical optimizer throughout the evolutionary search. Recall that for every LLM-generated class \(\Pi_{\Theta}\), the optimizer searches over \(\Theta\) to obtain a low-cost policy \(\widehat\pi\); this optimized policy, together with its parameter vector and empirical cost, then enters survival selection and may subsequently serve as a parent. Optimization can therefore affect \algo{} through two distinct channels. First, it can improve a generated class directly by calibrating its parameters. Second, because the optimized policy is used to evaluate the class and guide subsequent LLM proposals, it can change the evolutionary search path itself. In particular, a useful functional form with poorly calibrated LLM-supplied parameters may appear unattractive without optimization and be discarded before it can influence later generations. This section separates these two effects and asks whether optimization is valuable only for parameter calibration or also for guiding policy-class search.

\textbf{Experimental comparison.}
We compare two variants of \algo{} while holding the LLM backbone and the remaining search procedure fixed. The \emph{optimizer-throughout} variant is the default implementation used in \S\S\ref{sec:main}--\ref{sec:generalization}: for every feasible LLM proposal \((\Pi_{\Theta},\bm\theta_0)\), the external optimizer searches within \(\Theta\), and the resulting optimized policy is evaluated and used for selection and feedback. In the \emph{no-optimizer} variant, this within-class optimization step is removed. The policy instantiated at the LLM-supplied parameter vector \(\bm\theta_0\) is evaluated directly, and its functional form, raw parameter values, and empirical cost provide the feedback for subsequent generations. Thus, the two variants differ precisely in whether policy classes are evaluated after parameter optimization or at their LLM-supplied initialization. Regarding the demand instances, we use a subset of instances introduced in \S\ref{sec:main}: Normal with standard deviation $30$, Poisson, and Exponential, with a fixed $L = 6$ and cost parameters $(h, p) = (1, 2)$. We use the cost-reduction metric defined in \S\ref{sec:main}. Because the no-optimizer search improves more slowly, we allow it to run for \(20\) generations, twice the default \(10\)-generation budget of the optimizer-throughout variant. This gives the no-optimizer variant additional opportunity to compensate for the absence of numerical parameter search.

\begin{figure}[h]
    \centering
    \caption{Evolution of average cost reduction with and without external parameter optimization.}
    \includegraphics[
        width=\linewidth,
        trim=0 5mm 0 0,
        clip
    ]{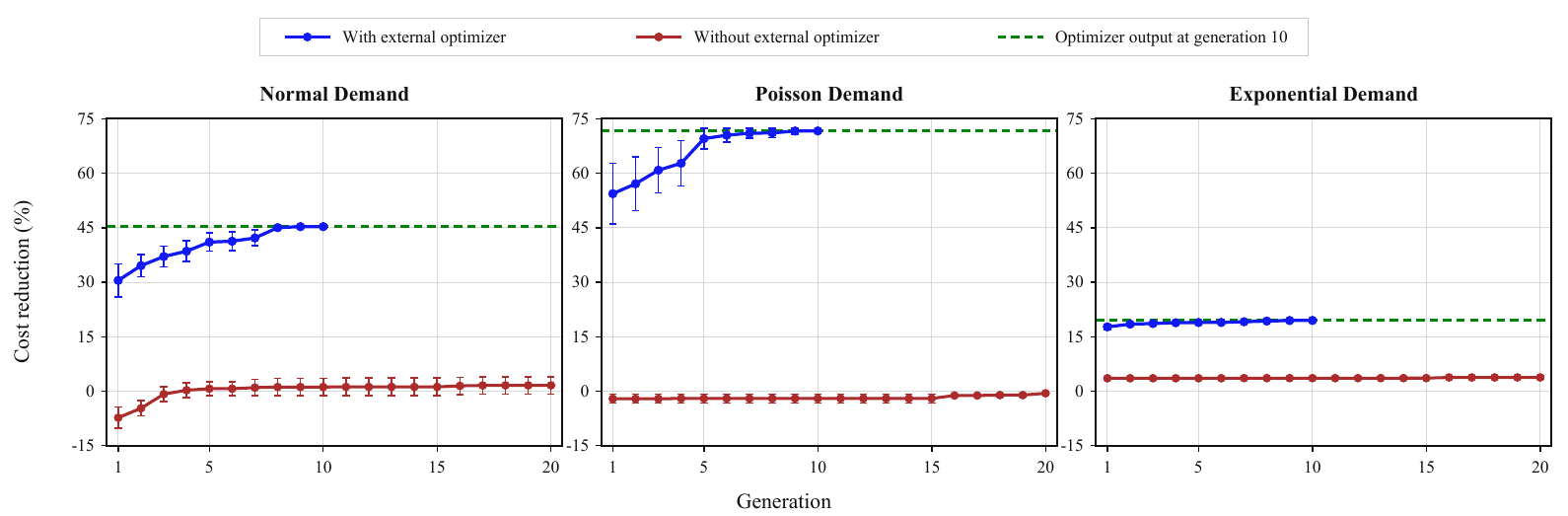}
    \label{fig:external_average_abs}
    \vspace{-30pt}
\end{figure}
\textbf{Performance level and search trajectory.}
Figure~\ref{fig:external_average_abs} compares the evolutionary trajectories under the two variants using DeepSeek V3, grouped by the Poisson, Normal, and Exponential demand environments. In all three panels, the horizontal axis denotes generation and the vertical axis reports average cost reduction relative to the optimized base-stock policy. The dotted line with diamonds corresponds to optimizer-throughout, while the solid line with circles corresponds to no-optimizer. 
As shown across the three distribution types, the difference is large from the start and persists throughout the search. With optimization, the average cost reduction is already positive in generation \(1\) and continues to rise by generation \(10\). Without optimization, the initial LLM-generated policies perform substantially worse than the optimized base-stock benchmark, and the search requires several generations merely to reach positive average improvement. Additional generations help, but do not close the gap: even after \(20\) generations, the no-optimizer variant is still well below the optimizer-throughout result after \(10\) generations. Thus, optimization does more than provide a small local refinement to LLM-generated parameter values. It places the evolutionary search on a substantially stronger performance trajectory from its earliest generations, and doubling the number of LLM search generations does not compensate for removing the numerical optimizer.


\begin{table}[h]
    \centering
    \caption{Convergence speed with and without external optimization.}
    \label{tab:external_speed}
    \begin{tabular*}{\linewidth}{@{\extracolsep{\fill}}lcccc@{}}
        \toprule
        & Median & Mean & Interquartile range & Range \\
        \midrule
        With optimizer    & 7.5  & 6.8  & [5,9]  & [1,10] \\
        Without optimizer & 15.0 & 11.6 & [4,18] & [1,20] \\
        \bottomrule
    \end{tabular*}
\end{table}

\textbf{Convergence speed.}
Table~\ref{tab:external_speed} examines whether optimization also accelerates convergence. For each run, we record the first generation at which the incumbent reaches within \(1\%\) of that variant's own final performance. Because this target is defined separately for each variant, the measure abstracts from differences in final performance and captures only how quickly each search approaches its own eventual level across all instances. With optimization, the median convergence time is \(7.5\) generations, compared with \(15.0\) generations without optimization; the corresponding means are \(6.8\) and \(11.6\) generations. The interquartile range is also substantially narrower with optimization, \([5,9]\) versus \([4,18]\). Thus, the external optimizer not only improves the quality of the policies discovered but also allows the evolutionary search to approach its eventual performance substantially faster and more consistently.


\textbf{Separating parameter calibration from search guidance.} The preceding comparisons establish that optimizer-throughout performs better and converges faster, but they do not yet identify why. One possibility is purely parametric: perhaps the no-optimizer search discovers equally useful policy classes, but performs poorly because the LLM assigns bad parameter values to those classes. If so, taking the class underlying the final no-optimizer policy and optimizing its parameters \emph{after} the evolutionary search should recover most of the performance gap. The alternative is a search-guidance effect: evaluating poorly calibrated policies during the search changes which classes survive, which policies become parents, and consequently which structural motifs the LLM explores in later generations. In that case, optimizing only the final no-optimizer class cannot recover the classes that were discarded or never reached because earlier feedback was less informative.

To distinguish these two effects, for each no-optimizer run we take the policy class underlying its final retained policy and optimize its parameters once after the evolutionary search is complete. Let \(\pi_k^{\mathrm{with}}\) denote the policy returned by optimizer-throughout, \(\pi_k^{\mathrm{no}}\) the policy returned by the no-optimizer search, and \(\pi_k^{\mathrm{post}}\) the policy obtained by applying final-step parameter optimization to the class underlying \(\pi_k^{\mathrm{no}}\). We define the total performance gap as the difference between the no-optimizer and optimizer-throughout policies, normalized by the optimized base-stock cost:
\[
G_k^{\mathrm{total}}:=100\frac{\widehat J_k(\pi_k^{\mathrm{no}})-\widehat J_k(\pi_k^{\mathrm{with}})}{\widehat J_k(\pi_{k,\mathrm B})}.
\]
Thus, \(G_k^{\mathrm{total}}\) measures how much performance is lost when optimization is removed from the evolutionary search. We then separate this loss into two components. The first, \(G_k^{\mathrm{param}}\), is the portion recovered by optimizing the parameters of the final class discovered without optimization; the second, \(G_k^{\mathrm{search}}\), is the gap that remains after this final-step optimization:
\[
G_k^{\mathrm{param}}:=100\frac{\widehat J_k(\pi_k^{\mathrm{no}})-\widehat J_k(\pi_k^{\mathrm{post}})}{\widehat J_k(\pi_{k,\mathrm B})}, \qquad G_k^{\mathrm{search}}:=100\frac{\widehat J_k(\pi_k^{\mathrm{post}})-\widehat J_k(\pi_k^{\mathrm{with}})}{\widehat J_k(\pi_{k,\mathrm B})}.
\]
By construction, \(G_k^{\mathrm{total}}=G_k^{\mathrm{param}}+G_k^{\mathrm{search}}\). A large \(G_k^{\mathrm{param}}\) would indicate that the benefit of optimization comes mainly from better parameter calibration within the final discovered class. In contrast, a large \(G_k^{\mathrm{search}}\) indicates that final-step calibration cannot recover most of the advantage of optimizer-throughout, consistent with optimization also improving the policy classes reached through the evolutionary search.

\begin{table}[t]
    \centering
    \caption{Decomposition of the test-performance gap between optimizer-throughout and no-optimizer search.}
    \label{tab:external_final_step_tuning}
    \begin{tabular*}{\linewidth}{@{\extracolsep{\fill}}lccc@{}}
        \toprule
        Demand & Total gap & Gap closed by final-step optimization & Remaining gap \\
        \midrule
        Poisson     & 72.9 & 1.1 & 71.8 \\
        Exponential & 16.1 & 0.0 & 16.1 \\
        Normal      & 43.8 & 3.1 & 40.7 \\
        \bottomrule
    \end{tabular*}
\end{table}

Table~\ref{tab:external_final_step_tuning} shows that final-step parameter optimization closes only a small fraction of the performance gap. For Poisson demand, only \(1.1\) percentage points of the \(72.9\)-point gap are recovered; for Exponential demand, essentially none of the \(16.1\)-point gap is recovered; and for Normal demand, only \(3.1\) of \(43.8\) percentage points are recovered. Thus, the large majority of the advantage from optimizer-throughout remains even after the final no-optimizer class is numerically calibrated. This decomposition provides the clearest evidence that the value of optimization extends beyond parameter tuning. By evaluating each proposed class after calibration, the optimizer provides more informative feedback for survival and parent selection throughout the evolutionary process. A promising class with poor LLM-supplied parameters can therefore remain in the search and influence subsequent proposals. Final-step optimization can improve the parameters of the class that eventually survives the no-optimizer search, but it cannot recover promising classes that were discarded earlier or the different evolutionary path that optimization-guided feedback would have produced.\looseness=-1

Overall, the experiments identify three roles for the external optimizer in \algo{}. It improves within-class parameter calibration, accelerates convergence, and, most importantly, improves the feedback that guides policy-class search. The last effect explains why simply adding optimization after the evolutionary search recovers only a small portion of the performance gap. It also reinforces the interpretation of \S\ref{sec:generalization}: the discovered functional forms can transfer across environments, but their parameters must be recalibrated to realize their value in a new environment. The next section examines whether these findings are robust across different LLM backbones.

\section{LLM Backbone Ablation}\label{sec:diff-LLMs}

The previous sections evaluate \algo{} using DeepSeek V3 as the LLM backbone. We now examine how sensitive the search is to this choice by comparing six backbones: GPT-5 Nano, Gemini 2.5 Flash-Lite, Grok 4.1 Fast Non-Reasoning, GPT-5 Mini, DeepSeek V3, and Gemini 3 Flash. We use the same demand instances as in \S\ref{sec:external}. Our goal is not to rank these models as general-purpose LLMs, but to assess their effectiveness as policy-class generators within \algo{}. A stronger backbone should propose useful executable policy structures and respond effectively to evolutionary feedback. We compare the resulting searches using mean cost reduction relative to the optimized base-stock benchmark.

\begin{figure}[h]
    \centering
    \caption{Mean cost reduction by \algo{} across generations for different LLM backbones.}
    \includegraphics[width=\textwidth]{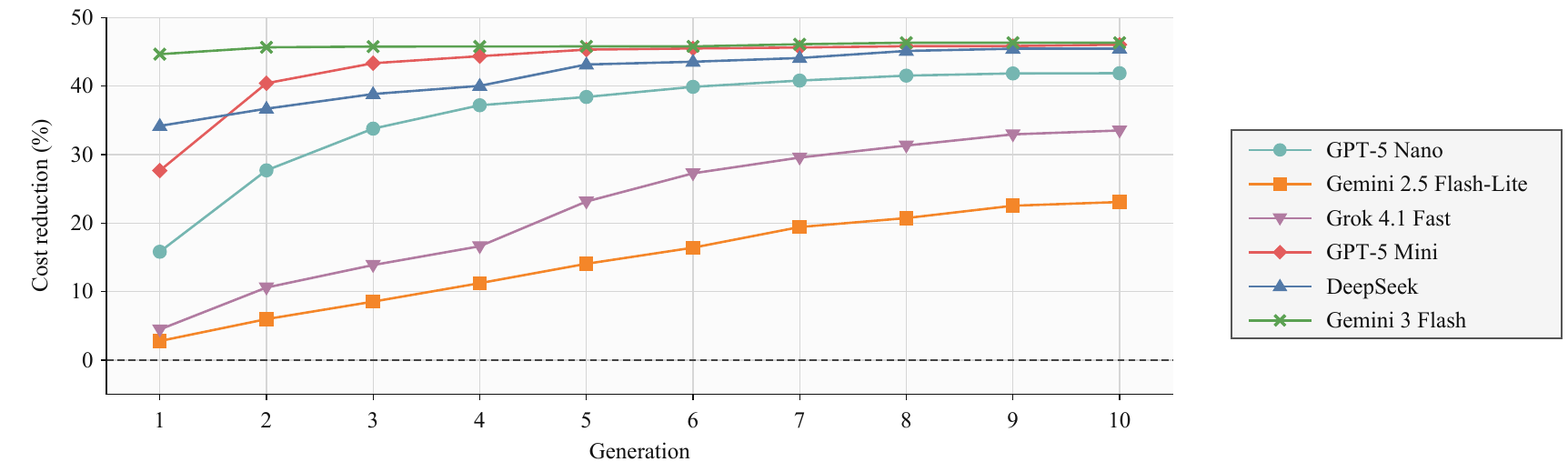}
    \label{fig:diff_llm_mean_performance}\vspace{-32pt}
\end{figure}
\textbf{Impact of the LLM backbone.}
Figure~\ref{fig:diff_llm_mean_performance} reports the mean cost reduction achieved by \algo{} across generations for each LLM backbone, with the external optimizer enabled. The horizontal axis denotes generation, and the vertical axis reports mean cost reduction relative to the optimized base-stock policy; the dashed horizontal line at zero marks the base-stock benchmark. Performance varies substantially across backbones. Gemini 3 Flash is strongest from the outset, achieving about \(45\%\) mean cost reduction in generation~1 and remaining around \(46\%\) thereafter. GPT-5 Mini rises from roughly \(28\%\) in generation~1 to above \(40\%\) by generation~2 and approaches \(46\%\) by generation~10. DeepSeek V3 also performs strongly, rising from about 34\% in generation 1 to approximately 45\% by generation 10. GPT-5 Nano improves more gradually, from about \(16\%\) to \(42\%\). Grok 4.1 Fast Non-Reasoning improves steadily, reaching approximately \(34\%\) by the final generation. Gemini 2.5 Flash-Lite is the weakest backbone in this experiment, although its mean cost reduction still increases from about \(3\%\) in generation~1 to \(23\%\) in generation~10. 

The key insight is that the choice of LLM backbone materially affects both the speed and the level of improvement achieved by \algo{}. Because the same external optimizer is used across backbones, these differences point to variation in the quality of the policy classes supplied by the LLM. Stronger backbones such as Gemini 3 Flash, GPT-5 Mini, and Deepseek V3 generate high-performing, optimizable structures early, whereas others require more generations or reach lower performance levels. At the same time, the steady improvement of GPT-5 Nano and Grok 4.1 Fast Non-Reasoning shows that evolutionary feedback can progressively improve the structural search even when the strongest policy classes are not identified immediately.

\begin{figure}[t]
    \centering
    \caption{Cost reduction across LLM backbones and generations with and without the external optimizer.}
    \includegraphics[width=\textwidth]{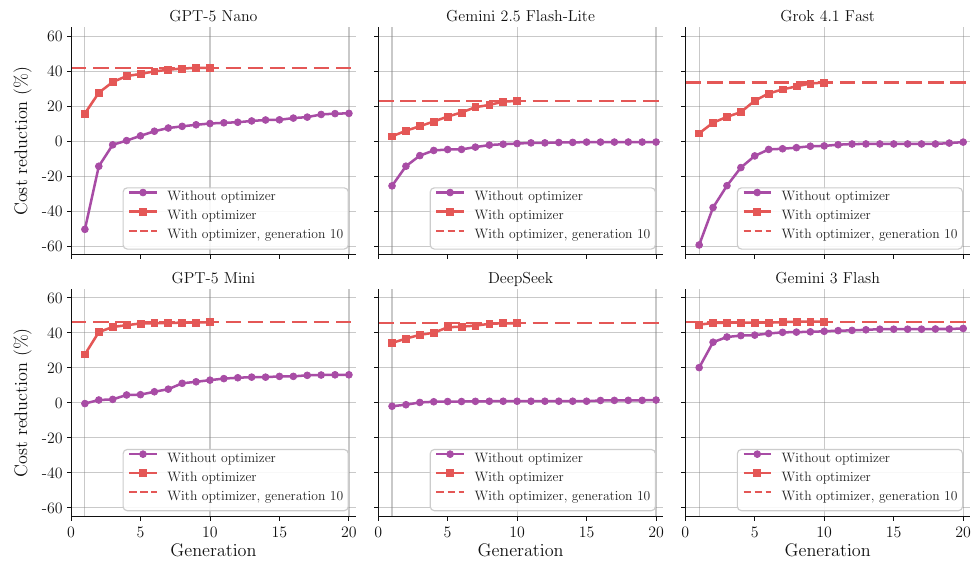}
    \label{fig:diff_llm_optimizer_ablation_boxplots}\vspace{-30pt}
\end{figure}

\textbf{Interaction between the LLM backbone and optimization.}
Figure~\ref{fig:diff_llm_optimizer_ablation_boxplots} examines how the value of the external optimizer varies across LLM backbones. Each panel corresponds to one backbone, with generation on the horizontal axis and mean cost reduction relative to the optimized base-stock policy on the vertical axis. The red squares show \algo{} with the optimizer over the default \(10\) generations, while the purple circles show the no-optimizer variant, which is allowed to continue for \(20\) generations. The dashed horizontal line marks the cost reduction attained by the optimizer-enabled variant at generation~10. Thus, the figure shows both the immediate performance effect of optimization and whether additional LLM generations can compensate for its absence.

Two patterns stand out. First, optimization improves performance substantially for every backbone and generally does so much earlier in the search. For GPT-5 Nano, mean cost reduction reaches about \(42\%\) with optimization by generation~10, compared with only about \(16\%\) without optimization after \(20\) generations. The corresponding comparison is approximately \(23\%\) versus \(0\%\) for Gemini 2.5 Flash-Lite, \(34\%\) versus \(0\%\) for Grok 4.1 Fast Non-Reasoning, \(46\%\) versus \(16\%\) for GPT-5 Mini, and \(45\%\) versus \(2\%\) for DeepSeek V3. In each of these cases, doubling the number of generations does not compensate for removing the optimizer. Second, the magnitude of this effect depends strongly on the backbone. Gemini 3 Flash is the notable exception: even without optimization, its mean cost reduction rises from about \(20\%\) in generation~1 to above \(40\%\) by later generations, approaching the roughly \(46\%\) attained with optimization. Thus, Gemini 3 Flash appears to supply both strong policy structures and relatively effective initial parameter values, whereas for backbones such as GPT-5 Mini and GPT-5 Nano, optimization is especially important for realizing the value of the generated classes.\looseness=-1

Together, Figures~\ref{fig:diff_llm_mean_performance} and~\ref{fig:diff_llm_optimizer_ablation_boxplots} clarify the complementary roles of the LLM and the optimizer in \algo{}. The LLM determines which policy classes enter the search, while the optimizer reveals how well those classes can perform after parameter calibration. Optimization, therefore, improves every backbone, but it cannot fully eliminate differences in structural-search quality across LLMs.

Moreover, Table~\ref{tab:llm_backbone_mean_sd}, which complements Figure~\ref{fig:diff_llm_optimizer_ablation_boxplots}, reports the generation-by-generation cost reductions with and without the optimizer, together with their standard deviations for each backbone. Table~\ref{tab:llm_backbone_test_convergence} provides a corresponding out-of-sample comparison at the termination of each search---generation~20 without the optimizer and generation~10 with the optimizer---reporting the mean, standard deviation, minimum, and maximum test-set cost reduction for each backbone. The test-set results exhibit the same broad performance patterns observed in the in-sample trajectories.

\begin{table}[h]
    \centering
    \caption{Observed LLM expense per repeat across backbones.}
    \label{tab:llm_generation_cost}

    \resizebox{\linewidth}{!}{%
    \setlength{\tabcolsep}{10pt}
    \begin{tabular}{lcccccc}
        \toprule
        LLM backbone &
        \shortstack[c]{Gemini 2.5\\Flash-Lite} &
        \shortstack[c]{GPT-5\\Nano} &
        \shortstack[c]{Grok 4.1 Fast\\Non-Reasoning} &
        \shortstack[c]{DeepSeek V3\\0324} &
        \shortstack[c]{GPT-5\\Mini} &
        \shortstack[c]{Gemini 3\\Flash} \\
        \midrule
        Expense (USD) &
        \$0.087 &
        \$0.101 &
        \$0.192 &
        \$0.243 &
        \$0.299 &
        \$0.505 \\
        \bottomrule
    \end{tabular}%
    }

    \vspace{3pt}
    \begin{minipage}{0.97\linewidth}
        \footnotesize
        \emph{Notes:} The reported values are the observed LLM expenses for one repeat under the common experimental configuration in \S\ref{sec:main}. They should therefore be interpreted as experiment-specific usage expenses rather than standalone model prices.
    \end{minipage}
\end{table}

\textbf{Expense--performance trade-off.}
The LLM backbones also differ substantially in their observed usage expenses. Table~\ref{tab:llm_generation_cost} shows that the expense of one repeat ranges from approximately \$0.087 for Gemini 2.5 Flash-Lite to \$0.505 for Gemini 3 Flash, a nearly sixfold difference. Despite this variation, all six backbones produce positive mean cost reductions relative to the optimized base-stock benchmark when combined with external optimization. Search performance, however, is not determined by LLM expense alone. The more expensive backbones, particularly Gemini 3 Flash, GPT-5 Mini, and Deepseek V3 tend to identify strong policy classes quickly and attain large cost reductions within relatively few generations. Yet less expensive backbones can also perform well. GPT-5 Nano, for example, has an observed expense of only \$0.101 per repeat but eventually achieves mean cost reductions above \(40\%\). Grok 4.1 Fast Non-Reasoning occupies an intermediate expense--performance region, while Gemini 2.5 Flash-Lite is the least expensive but also produces the smallest improvements.

These results show that the choice of LLM backbone affects both the quality and speed of policy-class search. Stronger backbones tend to identify high-performing policy classes earlier, although less expensive models can also perform well when combined with external optimization. More importantly, the external optimizer improves performance across all six backbones, but does not eliminate differences in structural-search quality. Thus, \algo{} does not depend on a particular LLM backbone, while improvements in LLM capability can directly translate into better and faster policy discovery.


\section{Conclusion}\label{sec:conclusion}

    In this paper, we study whether LLMs, numerical optimization, and data-driven evaluation can be combined to automate the design of inventory policy classes. We propose \algo{}, an evolutionary framework in which an LLM generates parameterized replenishment-policy classes, a numerical optimizer searches for strong policies within each class, and the resulting optimized policies and their empirical performance guide subsequent generations of structural search. Using lost-sales inventory systems with positive lead times as a test bed, we find that \algo{} discovers policy classes that substantially outperform the optimized base-stock benchmark. The resulting policies are also interpretable: rather than behaving as opaque black-box rules, they repeatedly combine recognizable inventory-control motifs, including richer representations of pipeline inventory, partial adjustment, clipping, thresholding, and state-dependent replenishment.

    These discovered structures also generalize beyond the instances on which they are generated. After retaining their functional forms and feasible parameter spaces and reoptimizing their parameters for new environments, the capped, discounted-pipeline, and hybrid classes achieve mean cost reductions exceeding \(21\%\) across more than \(10{,}000\) target instances spanning different demand distributions, lead times, and cost ratios. Their relative performance varies systematically across operating regimes, indicating that different motif combinations are useful under different inventory conditions. Our ablation experiments further show that numerical optimization is central to the search process: it not only calibrates parameters and accelerates convergence, but also improves the feedback used to select and evolve policy classes. Consistent with this mechanism, applying optimization only after the search recovers only a small portion of the performance lost when optimization is removed from the evolutionary loop. Comparisons across six LLM backbones also show that the quality of structural search depends on the LLM, while optimization improves realized performance across all backbones.

    More broadly, our findings suggest that LLMs can complement domain-specific knowledge about effective policies. In particular, \algo{} can take such expert-designed policies as starting points and systematically build on them by adding, combining, and refining structural motifs, potentially producing stronger policy classes than the original domain-based designs. Future work could develop richer mechanisms for representing and recombining structural motifs and extend the framework to settings with nonstationary demand, stochastic lead times, capacity constraints, multi-item systems, and multi-echelon networks.


\bibliographystyle{ormsv080} 
\bibliography{references} 

\ECSwitch

\ECDisclaimer

\section{Prompt Template}
\label{app:prompt_template}

We report the prompt template used in the LLM-guided policy-generation step of \algo{}. The prompt is reproduced verbatim. Placeholders represented in double braces indicate implementation-time substitutions; in the experiments, they are replaced by instance-specific inputs, including the training demand trajectories, selected parent-policy code, optimized parameter values, and empirical cost statistics. When the parent set contains multiple policies, the parent-policy block is repeated with the corresponding code and statistics.

Time is indexed differently in the prompt than in the main paper. In the main paper, the \(T\)-period cost-incurring horizon is indexed by \(t=1,\ldots,T\). 
In the prompt, we prepend \(L\) zero-demand, zero-cost periods, so that each simulated trajectory spans \(L+T\) periods. 
We refer to the prepended periods \(t=1,\ldots,L\) as the ``planning phase'' and the remaining periods \(t=L+1,\ldots,L+T\) as the ``selling phase.'' 
The selling phase corresponds one-to-one to the \(T\) periods in the main paper. 
The planning phase is used solely to initialize the pipeline: orders placed during these \(L\) periods become arrivals during the first \(L\) periods of the selling horizon. 
Thus, the expanded indexing in the prompt makes the pre-horizon pipeline initialization explicit.
\begin{lstlisting}[style=promptstyle]
[BEGIN PROMPT]

Section 1. Problem Statement

We consider an inventory control problem for a single product over a finite
discrete horizon. The system has a selling horizon of T periods and a fixed,
deterministic delivery lead time L. An order placed at the beginning of period t
arrives at the beginning of period t+L.

There are two phases:

1. Planning Phase (t = 1, ..., L):
   No customer demand occurs. The manager may place orders, but no costs are
   incurred. In the historical data, these periods appear as zeros, which are
   placeholders indicating absence of demand.

2. Selling Phase (t = L+1, ..., L+T):
   Customer demand occurs and costs are incurred. The objective is to minimize
   total cost over these T selling periods.

At the beginning of each period t = 1, ..., L+T, the manager observes:
- On-hand inventory I_t >= 0.
- Pipeline vector Q_t = (q_{t,1}, q_{t,2}, ..., q_{t,L}),
  where q_{t,k} is the quantity scheduled to arrive at the beginning of
  period t+k-1.

We assume I_1 = 0.

At the start of period t, the order q_{t,1} arrives. Before demand, the
available inventory is I_t + q_{t,1}. The manager then places an order a_t >= 0,
which will arrive at period t+L.

After ordering, demand D_t is realized. Sales and lost sales are:
    S_t = min(I_t + q_{t,1}, D_t),
    lost_sales_t = max(0, D_t - I_t - q_{t,1}).

The next on-hand inventory is:
    I_{t+1} = max(0, I_t + q_{t,1} - D_t).

Pipeline orders shift forward and the new order enters the last slot:
    Q_{t+1} = (q_{t,2}, q_{t,3}, ..., q_{t,L}, a_t).

During the selling phase (t = L+1, ..., L+T), the manager incurs:
- holding cost h * max(0, I_t + q_{t,1} - D_t),
- lost-sales cost p * max(0, D_t - I_t - q_{t,1}).

Thus the period-t cost is:
    c(I_t, q_{t,1}, D_t)
    = h * max(0, I_t + q_{t,1} - D_t)
    + p * max(0, D_t - I_t - q_{t,1}).

A stationary policy pi maps each state (I_t, Q_t) to an order quantity:
    a_t = pi(I_t, Q_t).

We are given N historical demand trajectories:
    D^n = (D_1^n, ..., D_L^n, D_{L+1}^n, ..., D_{L+T}^n),
    n = 1, ..., N,
where D_1^n, ..., D_L^n are zeros indicating no demand in the planning phase,
and the remaining T entries are actual demands.

The objective is to find a stationary policy minimizing the average total cost:
    min_pi (1/N) * sum_{n=1}^N sum_{t=L+1}^{L+T}
    c(I_t^{pi,n}, q_{t,1}^{pi,n}, D_t^n).

The simulator evaluates a policy using the following procedure:

For each demand trajectory n = 1, ..., N:
    Set I_1 = 0
    Set Q_1 = (0, 0, ..., 0)  # length L

    For each period t = 1, ..., L+T:
        q_{t,1} = first element of Q_t
        a_t = policy(I_t, Q_t)
        D_t = D_t^n

        If t > L:
            Add c(I_t, q_{t,1}, D_t) to total cost

        I_{t+1} = max(0, I_t + q_{t,1} - D_t)
        Q_{t+1} = (q_{t,2}, q_{t,3}, ..., q_{t,L}, a_t)

Return the average total cost over the historical demand trajectories.


Section 2. Problem Parameters and Training Data

Problem parameters:
- Selling phase horizon: T = 50 periods
- Lead time: L = 6 periods
- Holding cost: h = 1 per unit per period
- Lost-sales cost: p = 2 per unit

Historical demand trajectories:
{{TRAINING_DEMAND_TRAJECTORIES}}

Each trajectory has length L+T. The first L entries are zeros corresponding to
the planning phase, and the final T entries are selling-phase demands.


Section 3. Main Task

Given the problem parameters and historical demand trajectories above, use the
historical parent policy and its corresponding cost statistics in Section 4 to
generate a modified version of the policy. The modified policy should aim to
achieve a lower average total cost on the historical demand trajectories.

Each policy is represented by Python code. The function must be named:

    compute_order_amount

The function accepts two inputs:
- on_hand_inventory (float): on-hand inventory I_t at the start of the period.
- pipeline_orders (list[float]): pipeline orders
  Q_t = (q_{t,1}, q_{t,2}, ..., q_{t,L}), indexed from oldest to newest.

The function returns one output:
- order_amount (int): order quantity a_t for this period, with order_amount >= 0.

The list pipeline_orders is a FIFO queue of length L:
- pipeline_orders[0] is q_{t,1}, the oldest pipeline order arriving at the
  beginning of the current period.
- pipeline_orders[-1] is q_{t,L}, the newest pipeline order.


Instructions for Generating the Modified Policy

Follow these requirements when generating the modified policy:

1. Mark optimizable parameters in the code using comments of the form:
       # OPT_PARAM: {"initial": value, "min": lower, "max": upper, "type": "float"}

   Example:
       base_stock = 500.0  # OPT_PARAM: {"initial": 500.0, "min": 300.0, "max": 800.0, "type": "float"}

2. The OPT_PARAM comment must appear on the same line as the parameter assignment.

3. Only mark parameters that are assigned within the function body.

4. Only mark parameters assigned with an equals sign.

5. Do not mark more than 10 optimizable parameters.

6. The policy must be strictly stationary:
   - The order amount may depend only on on_hand_inventory and pipeline_orders.
   - The order amount may not depend on time, step counters, episode index,
     demand trajectory index, or previous calls.
   - Do not use hidden memory, function attributes, cached state, global variables,
     module-level variables, or closures that store information across calls.

7. The function must always return a finite nonnegative integer order quantity.

8. After generating the modified implementation, provide a concise explanation
   for the changes inside double curly braces, like this:
       {{Your explanation here.}}

9. Do not provide any additional explanations outside the code and the double
   curly-brace explanation.


Section 4. Historical Parent Policy and Cost Statistics

Parent policy 1:

def compute_order_amount(on_hand_inventory, pipeline_orders):
    base_stock = 497.94  # OPT_PARAM: {"initial": 497.94, "min": 400, "max": 700, "type": "float"}
    safety_stock = 25.03  # OPT_PARAM: {"initial": 25.03, "min": 20, "max": 100, "type": "float"}
    demand_forecast = 86.14  # OPT_PARAM: {"initial": 86.14, "min": 80, "max": 110, "type": "float"}
    smoothing_factor = 0.05  # OPT_PARAM: {"initial": 0.05, "min": 0.05, "max": 0.30, "type": "float"}
    pipeline_weight = 0.50  # OPT_PARAM: {"initial": 0.50, "min": 0.50, "max": 1.00, "type": "float"}

    weighted_pipeline = 0.0
    for i, q in enumerate(pipeline_orders):
        weight = pipeline_weight ** (len(pipeline_orders) - i - 1)
        weighted_pipeline += q * weight

    net_inventory = on_hand_inventory + weighted_pipeline

    pipeline_sum = sum(pipeline_orders)
    if pipeline_sum > 0:
        pipeline_ratio = weighted_pipeline / pipeline_sum
        adjusted_safety = safety_stock * (1.0 + 0.5 * (1.0 - pipeline_ratio))
    else:
        adjusted_safety = safety_stock

    target_inventory = base_stock + adjusted_safety
    order_up_to = max(0.0, target_inventory - net_inventory)

    smoothed_order = (
        smoothing_factor * order_up_to
        + (1.0 - smoothing_factor) * demand_forecast
    )

    order_amount = max(0, int(round(smoothed_order)))
    return order_amount

Cost statistics of parent policy 1 on the historical demand trajectories:
- Average total cost per period: 14.76
- Standard deviation of total cost per period: 14.14
- Average holding cost per period: 5.67
- Average lost-sales cost per period: 9.10
- Holding-to-lost-sales ratio: 5.7 : 9.1
- Per-trajectory total cost:
    mean = 826.80
    standard deviation = 163.23
    range = (494.00, 1584.00)

Now generate one modified policy.

[END PROMPT]
\end{lstlisting}

\section{Additional Results for Section~\ref{sec:main}}\label{sec:main_apndx}

Table~\ref{tab:cr-by-instance-generation-compact} reports the mean and standard error of the cost reduction percentage across different problem instances and generations. Each problem instance occupies two rows: the first row reports the mean of $\mathrm{CR}_{k,r}^{(g)}$ over the ten repeats, and the parenthesized second row reports its standard error. In the instance names, \texttt{Exp} denotes Exponential demand; \texttt{L2}, \texttt{L4}, and \texttt{L6} denote the lead time; \texttt{CR2} and \texttt{CR5} denote the cost ratio $p/h=2$ and $p/h=5$; and \texttt{Normal10}, \texttt{Normal30}, and \texttt{Normal50} denote Normal demand with $\sigma=10,30,50$, respectively.  

\begin{table}[!ht]
\centering
\caption{Mean (standard error) of $\mathrm{CR}_{k,r}^{(g)}$ by problem instance and generation}
\label{tab:cr-by-instance-generation-compact}
\fontsize{9.0}{9.4}\selectfont
\setlength{\tabcolsep}{3.6pt}
\renewcommand{\arraystretch}{0.86}

\begin{tabular*}{0.92\textwidth}{@{\extracolsep{\fill}}l *{10}{r}@{}}
\toprule
\multicolumn{1}{c}{\textbf{Instance}} & \multicolumn{1}{c}{$g=1$} & \multicolumn{1}{c}{$g=2$} & \multicolumn{1}{c}{$g=3$} & \multicolumn{1}{c}{$g=4$} & \multicolumn{1}{c}{$g=5$} & \multicolumn{1}{c}{$g=6$} & \multicolumn{1}{c}{$g=7$} & \multicolumn{1}{c}{$g=8$} & \multicolumn{1}{c}{$g=9$} & \multicolumn{1}{c}{$g=10$} \\
\midrule
\multirow{2}{*}{\texttt{Poisson\_L2\_CR2}} & 10.98 & 16.68 & 20.09 & 22.74 & 25.02 & 25.79 & 26.19 & 28.41 & 28.50 & 28.53 \\[-1.55pt]
 & (3.51) & (3.19) & (2.68) & (2.54) & (2.53) & (2.64) & (2.69) & (1.08) & (1.10) & (1.11) \\[0.25pt]
\multirow{2}{*}{\texttt{Poisson\_L2\_CR5}} & 7.52 & 11.58 & 12.53 & 14.71 & 16.34 & 16.91 & 17.36 & 17.62 & 17.87 & 17.91 \\[-1.55pt]
 & (1.85) & (1.10) & (1.27) & (1.50) & (1.28) & (1.28) & (1.13) & (1.08) & (1.12) & (1.12) \\[0.25pt]
\multirow{2}{*}{\texttt{Poisson\_L4\_CR2}} & 24.37 & 41.15 & 47.30 & 48.88 & 52.86 & 54.01 & 55.02 & 55.38 & 55.86 & 56.16 \\[-1.55pt]
 & (6.47) & (4.84) & (3.58) & (3.58) & (1.64) & (1.27) & (0.80) & (0.72) & (0.67) & (0.60) \\[0.25pt]
\multirow{2}{*}{\texttt{Poisson\_L4\_CR5}} & 17.95 & 31.67 & 36.82 & 40.28 & 42.00 & 43.55 & 44.69 & 45.55 & 45.68 & 45.88 \\[-1.55pt]
 & (5.59) & (3.11) & (2.76) & (2.13) & (2.10) & (1.67) & (1.06) & (0.73) & (0.79) & (0.79) \\[0.25pt]
\multirow{2}{*}{\texttt{Poisson\_L6\_CR2}} & 54.26 & 56.99 & 60.74 & 62.65 & 69.46 & 70.41 & 70.95 & 71.02 & 71.57 & 71.58 \\[-1.55pt]
 & (8.40) & (7.40) & (6.31) & (6.27) & (2.88) & (1.94) & (1.42) & (1.35) & (0.82) & (0.82) \\[0.25pt]
\multirow{2}{*}{\texttt{Poisson\_L6\_CR5}} & 44.99 & 55.24 & 57.51 & 60.58 & 63.61 & 64.75 & 64.85 & 64.87 & 64.87 & 64.93 \\[-1.55pt]
 & (8.60) & (6.54) & (6.23) & (3.92) & (1.18) & (0.32) & (0.31) & (0.29) & (0.29) & (0.30) \\[1.35pt]
\multirow{2}{*}{\texttt{Normal10\_L2\_CR2}} & 11.00 & 18.74 & 24.72 & 25.25 & 25.47 & 25.80 & 25.97 & 28.85 & 28.99 & 29.03 \\[-1.55pt]
 & (3.85) & (4.01) & (2.86) & (2.93) & (2.91) & (2.90) & (2.93) & (1.10) & (1.11) & (1.12) \\[0.25pt]
\multirow{2}{*}{\texttt{Normal10\_L2\_CR5}} & 5.12 & 12.35 & 15.49 & 16.66 & 17.05 & 17.52 & 17.55 & 18.42 & 18.44 & 18.51 \\[-1.55pt]
 & (2.12) & (2.18) & (1.95) & (1.71) & (1.67) & (1.68) & (1.68) & (1.82) & (1.81) & (1.84) \\[0.25pt]
\multirow{2}{*}{\texttt{Normal10\_L4\_CR2}} & 45.64 & 52.05 & 52.84 & 53.08 & 53.25 & 53.38 & 54.33 & 55.05 & 55.07 & 55.18 \\[-1.55pt]
 & (3.43) & (1.74) & (1.63) & (1.57) & (1.56) & (1.58) & (0.99) & (0.61) & (0.62) & (0.58) \\[0.25pt]
\multirow{2}{*}{\texttt{Normal10\_L4\_CR5}} & 6.02 & 12.98 & 18.36 & 22.54 & 27.41 & 34.08 & 34.47 & 35.02 & 35.21 & 35.36 \\[-1.55pt]
 & (3.98) & (4.93) & (5.66) & (5.75) & (5.08) & (4.44) & (4.49) & (4.40) & (4.37) & (4.27) \\[0.25pt]
\multirow{2}{*}{\texttt{Normal10\_L6\_CR2}} & 47.54 & 64.66 & 68.00 & 70.74 & 70.74 & 71.92 & 72.32 & 72.32 & 72.65 & 72.70 \\[-1.55pt]
 & (8.78) & (4.22) & (2.52) & (1.83) & (1.83) & (0.88) & (0.78) & (0.78) & (0.57) & (0.55) \\[0.25pt]
\multirow{2}{*}{\texttt{Normal10\_L6\_CR5}} & 34.13 & 46.71 & 49.98 & 57.01 & 58.88 & 59.69 & 59.71 & 60.68 & 60.96 & 61.94 \\[-1.55pt]
 & (8.94) & (6.93) & (6.28) & (2.82) & (2.39) & (2.41) & (2.41) & (1.46) & (1.22) & (0.43) \\[1.35pt]
\multirow{2}{*}{\texttt{Normal30\_L2\_CR2}} & 5.28 & 8.28 & 11.00 & 11.29 & 11.33 & 11.51 & 11.59 & 11.74 & 11.81 & 11.85 \\[-1.55pt]
 & (1.71) & (1.25) & (0.42) & (0.43) & (0.43) & (0.45) & (0.47) & (0.45) & (0.42) & (0.42) \\[0.25pt]
\multirow{2}{*}{\texttt{Normal30\_L2\_CR5}} & 3.71 & 4.96 & 6.58 & 6.70 & 7.24 & 7.75 & 7.95 & 8.30 & 8.50 & 8.68 \\[-1.55pt]
 & (1.28) & (1.07) & (0.99) & (0.99) & (1.02) & (0.66) & (0.66) & (0.54) & (0.53) & (0.56) \\[0.25pt]
\multirow{2}{*}{\texttt{Normal30\_L4\_CR2}} & 20.98 & 29.21 & 31.77 & 32.51 & 32.61 & 33.08 & 33.21 & 33.32 & 33.50 & 33.52 \\[-1.55pt]
 & (4.58) & (1.90) & (0.81) & (0.46) & (0.48) & (0.32) & (0.27) & (0.27) & (0.24) & (0.23) \\[0.25pt]
\multirow{2}{*}{\texttt{Normal30\_L4\_CR5}} & 18.61 & 20.62 & 22.24 & 23.88 & 24.17 & 24.44 & 24.63 & 24.74 & 24.80 & 24.86 \\[-1.55pt]
 & (2.72) & (1.72) & (1.41) & (0.64) & (0.45) & (0.29) & (0.22) & (0.15) & (0.16) & (0.15) \\[0.25pt]
\multirow{2}{*}{\texttt{Normal30\_L6\_CR2}} & 29.88 & 33.98 & 36.51 & 38.00 & 40.52 & 40.77 & 41.67 & 44.54 & 44.82 & 44.83 \\[-1.55pt]
 & (4.57) & (3.05) & (2.84) & (2.88) & (2.54) & (2.54) & (2.19) & (0.53) & (0.52) & (0.52) \\[0.25pt]
\multirow{2}{*}{\texttt{Normal30\_L6\_CR5}} & 12.66 & 25.48 & 28.92 & 31.65 & 32.94 & 33.08 & 33.55 & 34.03 & 34.22 & 34.45 \\[-1.55pt]
 & (4.15) & (3.87) & (2.64) & (1.57) & (1.01) & (1.03) & (0.97) & (0.72) & (0.75) & (0.66) \\[1.35pt]
\multirow{2}{*}{\texttt{Normal50\_L2\_CR2}} & 6.77 & 8.54 & 9.40 & 9.72 & 11.53 & 11.72 & 11.84 & 11.90 & 11.97 & 12.03 \\[-1.55pt]
 & (1.54) & (1.20) & (1.22) & (1.13) & (0.22) & (0.19) & (0.16) & (0.18) & (0.16) & (0.12) \\[0.25pt]
\multirow{2}{*}{\texttt{Normal50\_L2\_CR5}} & 3.50 & 4.55 & 4.97 & 5.31 & 5.87 & 5.92 & 5.92 & 6.06 & 6.08 & 6.22 \\[-1.55pt]
 & (0.88) & (0.44) & (0.37) & (0.36) & (0.37) & (0.37) & (0.37) & (0.37) & (0.37) & (0.33) \\[0.25pt]
\multirow{2}{*}{\texttt{Normal50\_L4\_CR2}} & 11.00 & 16.75 & 17.23 & 17.42 & 17.83 & 17.93 & 18.05 & 18.54 & 18.95 & 19.01 \\[-1.55pt]
 & (2.39) & (0.54) & (0.58) & (0.60) & (0.52) & (0.51) & (0.50) & (0.40) & (0.38) & (0.39) \\[0.25pt]
\multirow{2}{*}{\texttt{Normal50\_L4\_CR5}} & 9.89 & 10.50 & 10.96 & 11.81 & 11.88 & 13.54 & 13.72 & 13.86 & 14.05 & 14.05 \\[-1.55pt]
 & (1.74) & (1.69) & (1.77) & (1.63) & (1.64) & (0.66) & (0.69) & (0.69) & (0.71) & (0.71) \\[0.25pt]
\multirow{2}{*}{\texttt{Normal50\_L6\_CR2}} & 20.11 & 24.41 & 28.31 & 28.80 & 29.29 & 29.75 & 29.79 & 30.39 & 30.49 & 30.83 \\[-1.55pt]
 & (3.75) & (3.50) & (1.56) & (1.13) & (0.76) & (0.65) & (0.66) & (0.33) & (0.27) & (0.19) \\[0.25pt]
\multirow{2}{*}{\texttt{Normal50\_L6\_CR5}} & 14.75 & 17.72 & 18.31 & 20.21 & 21.80 & 22.54 & 24.16 & 24.59 & 24.73 & 24.86 \\[-1.55pt]
 & (2.92) & (2.33) & (2.35) & (2.24) & (1.43) & (1.43) & (0.43) & (0.27) & (0.19) & (0.20) \\[1.35pt]
\multirow{2}{*}{\texttt{Exp\_L2\_CR2}} & 5.77 & 7.30 & 8.36 & 8.47 & 8.54 & 8.60 & 8.63 & 8.66 & 8.76 & 8.83 \\[-1.55pt]
 & (0.84) & (0.59) & (0.13) & (0.12) & (0.12) & (0.11) & (0.12) & (0.11) & (0.10) & (0.11) \\[0.25pt]
\multirow{2}{*}{\texttt{Exp\_L2\_CR5}} & 6.54 & 7.79 & 8.16 & 8.19 & 8.35 & 8.94 & 9.00 & 9.00 & 9.01 & 9.01 \\[-1.55pt]
 & (0.81) & (0.59) & (0.58) & (0.57) & (0.56) & (0.06) & (0.03) & (0.03) & (0.03) & (0.03) \\[0.25pt]
\multirow{2}{*}{\texttt{Exp\_L4\_CR2}} & 10.12 & 11.23 & 11.33 & 11.50 & 11.59 & 11.60 & 11.73 & 11.84 & 11.90 & 12.08 \\[-1.55pt]
 & (1.14) & (0.23) & (0.24) & (0.25) & (0.25) & (0.25) & (0.16) & (0.14) & (0.16) & (0.08) \\[0.25pt]
\multirow{2}{*}{\texttt{Exp\_L4\_CR5}} & 9.42 & 9.51 & 9.87 & 10.51 & 10.58 & 10.63 & 10.70 & 10.76 & 10.83 & 10.83 \\[-1.55pt]
 & (0.72) & (0.72) & (0.65) & (0.24) & (0.25) & (0.27) & (0.27) & (0.25) & (0.26) & (0.26) \\[0.25pt]
\multirow{2}{*}{\texttt{Exp\_L6\_CR2}} & 17.13 & 17.89 & 18.06 & 18.27 & 18.33 & 18.35 & 18.55 & 18.73 & 18.89 & 18.91 \\[-1.55pt]
 & (0.74) & (0.39) & (0.38) & (0.36) & (0.34) & (0.33) & (0.26) & (0.16) & (0.12) & (0.12) \\[0.25pt]
\multirow{2}{*}{\texttt{Exp\_L6\_CR5}} & 10.02 & 13.49 & 14.51 & 14.94 & 14.97 & 15.15 & 15.35 & 15.41 & 15.67 & 15.73 \\[-1.55pt]
 & (1.57) & (1.03) & (0.52) & (0.49) & (0.49) & (0.51) & (0.44) & (0.46) & (0.33) & (0.34) \\[0.25pt]
\bottomrule
\end{tabular*}
 
\end{table}

The mean and standard error of cost reduction for a subset of problem instances $ \mathcal{K} \subset [K]$ are computed as
\[
\mathrm{MCR}^{(g)}(\mathcal{K}):=\frac{1}{|\mathcal{K}|}\sum_{ k \in \mathcal{K} R } \sum_{r=1}^R \mathrm{CR}_{k,r}^{(g)},\qquad
\mathrm{SE}^{(g)}(\mathcal{K}):= \frac{\mathrm{stdev}(\mathrm{CR}_{k,r}^{(g)} \ : \ k \in \mathcal{K},  r = 1, \dots, R)}{\sqrt{|\mathcal{K}| \cdot R}}.
\]

\section{Construction of the Robustness Demand Set}
\label{app:demand_generation}

This appendix describes the parameter grids and simulation procedures
used to construct the demand specifications in the cross-environment
robustness analysis. We consider 13 parametric demand families,
covering bounded and unbounded distributions, continuous and discrete
distributions, and demand environments with different degrees of
variability, skewness, and tail behavior.

\subsection{Trajectory Generation}

A demand specification is defined by a distribution family
\(\mathcal F\) and a vector of family-specific parameters.
For each specification, we generate integer-valued demand trajectories
of length \(T=50\). Demand observations from discrete distributions are
sampled directly from their integer-valued support. For a continuous
distribution, we first generate a continuous random variable
\(\widetilde D_t\) and then construct integer demand according to
\[
    D_t
    =
    \max\left\{
        0,\operatorname{round}(\widetilde D_t)
    \right\}.
\]
Thus, continuous draws are rounded to the nearest integer, and any
negative rounded values are replaced by zero.

For each demand specification, we generate \(100\) training
trajectories and \(1{,}000\) separate test trajectories. The training
trajectories are used to optimize the parameters of each policy class
within the corresponding environment, whereas the test trajectories
are reserved for out-of-sample performance evaluation.

Each demand specification is evaluated under four lead times, $L\in\{2,4,6,8\},$ and four lost-sales penalties, $p \in\{2,4,6,10\},$ with the unit holding cost fixed at \(h=1\). Consequently, each demand specification generates $4\times 4=16$ distinct inventory environments. 
The construction described below produces \(629\) distinct demand specifications. It therefore yields $629\times4\times4 = 10{,}064$ distinct inventory test environments.

\subsection{Distribution Families and Parameter Grids}

Table~\ref{tab:robustness_demand_grid} reports the complete parameter
grid used for each demand family. To avoid confusion with the
lost-sales penalty, \(q\) denotes a probability parameter in the
negative-binomial, binomial, and geometric distributions. For the
zero-based geometric distribution, the random variable records the
number of failures before the first success and therefore has support
\(\{0,1,2,\ldots\}\). The negative-binomial distribution is similarly
parameterized by the number of failures before \(r\) successes.

\begingroup

\makeatletter
\setlength{\LTcapwidth}{\textwidth}
\renewcommand{\LT@makecaption}[3]{%
  \LT@mcol\LT@cols c{%
    \hbox to\z@{\hss
      \parbox[t]{\LTcapwidth}{%
        \@maketablecaption{#1{#2}}{#3}%
      }%
    \hss}%
  }%
}
\makeatother

\footnotesize
\setlength{\tabcolsep}{4pt}
\setlength{\LTleft}{0pt}
\setlength{\LTright}{0pt}
\renewcommand{\arraystretch}{1.12}

\begin{longtable}{
    @{}
    P{0.20\textwidth}
    P{0.68\textwidth}
    N{0.07\textwidth}
    @{}
}
\caption[Parameter grids used to construct the robustness demand set.]{%
\phantomsection\label{tab:robustness_demand_grid}%
Parameter grids used to construct the robustness demand set.%
}
\\

\toprule
\textbf{Demand family}
&
\textbf{Parameter grid and restrictions}
&
\textbf{Count}
\\
\midrule
\endfirsthead

\multicolumn{3}{@{}l}{
    \footnotesize\textit{
        \tablename\ \thetable\ continued from the previous page
    }
}
\\
\addlinespace[2pt]

\toprule
\textbf{Demand family}
&
\textbf{Parameter grid and restrictions}
&
\textbf{Count}
\\
\midrule
\endhead

\midrule
\multicolumn{3}{r@{}}{
    \footnotesize\textit{Continued on the next page}
}
\\
\endfoot

\bottomrule
\endlastfoot


\multicolumn{3}{@{}l}{
    \textbf{\textit{(i) Bounded continuous distributions}}
}
\\
\addlinespace[2pt]

Scaled beta
&
\(\widetilde D=M X\), where
\(X\sim\operatorname{Beta}(\alpha,\beta)\);
\(\alpha,\beta\in\{0.5,1,2,5,10\}\);
and \(M\in\{20,50,100,200,500\}\).
The raw grid contains \(125\) combinations; two duplicate
moment specifications are removed.
&
123
\\
\addlinespace[3pt]

Continuous uniform
&
\(\widetilde D\sim\operatorname{Uniform}(a,b)\), where
\(a\in\{0,10,20,50\}\) and
\(b\in\{50,100,200,400,800\}\), retaining only combinations
satisfying \(b>a\).
&
19
\\
\addlinespace[3pt]

Triangular
&
\(\widetilde D\sim\operatorname{Triangular}(a,c,b)\), where
\(a\in\{0,5,10\}\),
\(c\in\{20,50,100,200\}\), and
\(b\in\{60,120,250,500,800\}\), retaining only combinations
satisfying \(a<c<b\).
&
51
\\
\addlinespace[6pt]


\multicolumn{3}{@{}l}{
    \textbf{\textit{(ii) Bounded discrete distributions}}
}
\\
\addlinespace[2pt]

Binomial
&
\(D\sim\operatorname{Binomial}(n,q)\), where
\(n\in\{20,50,100,200,400,800\}\) and
\(q\in\{0.05,0.1,0.2,0.3,0.5,0.7,0.9,0.95\}\).
&
48
\\
\addlinespace[3pt]

Discrete uniform
&
\(D\) is uniformly distributed on
\(\{\ell,\ell+1,\ldots,u\}\), where
\(\ell\in\{0,5,10,20,50\}\) and
\(u\in\{20,50,100,200,400,800\}\), retaining only combinations
satisfying \(u>\ell\).
&
27
\\
\addlinespace[6pt]


\multicolumn{3}{@{}l}{
    \textbf{\textit{(iii) Light- to moderate-tailed unbounded distributions}}
}
\\
\addlinespace[2pt]

Gamma
&
\(\widetilde D\sim\operatorname{Gamma}(k,\theta)\), where
\(k\in\{0.8,1,2,5,10,20\}\) is the shape parameter and
\(\theta\in\{2,5,10,20,40,80\}\) is the scale parameter.
&
36
\\
\addlinespace[3pt]

Geometric, zero-based
&
\(D\sim\operatorname{Geometric}_{0}(q)\), where
\(q\in\{0.03,0.05,0.08,0.1,0.15,0.2,0.3,0.4,0.6,0.8\}\).
The support is \(\{0,1,2,\ldots\}\).
&
10
\\
\addlinespace[3pt]

Negative binomial
&
\(D\sim\operatorname{NegBin}(r,q)\), where
\(r\in\{2,5,10,20,40\}\) and
\(q\in\{0.1,0.2,0.3,0.4,0.6,0.8\}\).
&
30
\\
\addlinespace[3pt]

Normal
&
\(\widetilde D\sim\mathcal N(\mu,\sigma^2)\), where
\(\mu\in\{20,50,80,100,150,200,300\}\) and
\(\sigma=f\mu\), with
\(f\in\{0.10,0.15,0.20,0.25,0.30\}\).
The specifications
\((\mu,\sigma)=(100,10)\) and \((100,30)\), already used in the
main experiments, are excluded.
&
33
\\
\addlinespace[3pt]

Weibull
&
\(\widetilde D\sim\operatorname{Weibull}(k,\lambda)\), where
\(k\in\{0.8,1.2,1.8,2.5,3.5,5.0\}\) is the shape parameter and
\(\lambda\in\{10,30,60,100,200,500\}\) is the scale parameter.
&
36
\\
\addlinespace[6pt]


\multicolumn{3}{@{}l}{
    \textbf{\textit{(iv) Heavy-tailed or highly skewed distributions}}
}
\\
\addlinespace[2pt]

Lognormal
&
\(\widetilde D\sim
\operatorname{Lognormal}(\mu_{\log},\sigma_{\log}^{2})\), where
\(\mu_{\log}\in\{1.5,2.0,2.5,3.0,3.5,4.0\}\) and
\(\sigma_{\log}\in\{0.3,0.5,0.7,1.0,1.3\}\).
These are the parameters of the underlying normal distribution.
&
30
\\
\addlinespace[3pt]

Pareto type I
&
\(\widetilde D\sim\operatorname{ParetoI}(\alpha,x_m)\), where
\(\alpha\in\{1.3,1.5,2.0,3.0,4.0,6.0\}\) is the shape parameter and
\(x_m\in\{1,5,10,20,50,100\}\) is the minimum support value.
&
36
\\
\addlinespace[3pt]

Zero-inflated negative binomial
&
Demand equals zero with probability
\(\pi_0\in\{0.1,0.3,0.5,0.7,0.85\}\).
Otherwise,
\(D\sim\operatorname{NegBin}(r,q)\), where
\(r\in\{2,5,10,20,40\}\) and
\(q\in\{0.1,0.2,0.3,0.4,0.6,0.8\}\).
&
150
\\
\addlinespace[4pt]

\midrule
\textbf{Total}
&
&
\textbf{629}
\\

\end{longtable}
\endgroup
We retain one representative from each duplicate pair, leaving
\(125-2=123\) distinct scaled-beta moment specifications. Together with
the specifications from the other 12 distribution families, this
produces the final set of \(629\) demand specifications.

\section{Test-Set Performance for the Value of Optimization}\label{sec:test_set_optimizer}

This appendix examines the impact of the optimizer on test-set performance for our AIPS algorithm. As mentioned in Section \ref{sec:external} of the main body, the test-set performance is very close to the training-set performance reported in the main body.

 In Table \ref{tab:final-generation-test-improvement}, the results without the external optimizer are evaluated at generation 20, whereas results with the external optimizer are evaluated at generation 10. These results are consistent with the training-set performance reported in the main body, showing that incorporating external optimizers can significantly improve the performance of the proposed AIPS algorithm.

\begin{table}[t]
  \centering
  \caption{Test-set percentage improvement over the optimal base-stock policy at the final generation. Each statistic is computed across 10 independent repeats.}
  \label{tab:final-generation-test-improvement}
  \begin{tabular}{lrrrr}
    \toprule
    Setting and demand distribution & Min (\%) & Max (\%) & Avg. (\%) & Std. dev. (pp) \\
    \midrule
    Without optimizer -- Normal      & -4.16 & 22.72 &  1.57 & 7.56 \\
    Without optimizer -- Exponential &  0.66 &  5.68 &  1.18 & 1.58 \\
    Without optimizer -- Poisson     & -1.07 & -0.13 & -0.79 & 0.45 \\
    \midrule
    With optimizer -- Normal         & 41.58 & 46.32 & 45.34 & 1.57 \\
    With optimizer -- Exponential    & 16.13 & 17.53 & 17.15 & 0.40 \\
    With optimizer -- Poisson        & 64.78 & 73.20 & 72.12 & 2.65 \\
    \bottomrule
  \end{tabular}
\end{table}

\begin{table}[t]
\centering
\caption{Cost reduction across LLM backbones with and without numerical parameter optimization.}
\label{tab:llm_backbone_mean_sd}

\resizebox{\textwidth}{!}{%
\begin{tabular}{c cc cc cc cc cc cc}
\toprule
& \multicolumn{2}{c}{GPT-5 Nano}
& \multicolumn{2}{c}{GPT-5 Mini}
& \multicolumn{2}{c}{DeepSeek}
& \multicolumn{2}{c}{Grok 4.1 Fast}
& \multicolumn{2}{c}{Gemini 2.5 Lite}
& \multicolumn{2}{c}{Gemini 3 Flash} \\
\cmidrule(lr){2-3}
\cmidrule(lr){4-5}
\cmidrule(lr){6-7}
\cmidrule(lr){8-9}
\cmidrule(lr){10-11}
\cmidrule(lr){12-13}
Generation
& Without & With
& Without & With
& Without & With
& Without & With
& Without & With
& Without & With \\
\midrule

1
& \shortstack{-50.4\\[-1pt]{\scriptsize (10.2)}}
& \shortstack{15.8\\[-1pt]{\scriptsize (3.0)}}
& \shortstack{-0.4\\[-1pt]{\scriptsize (0.8)}}
& \shortstack{27.7\\[-1pt]{\scriptsize (5.2)}}
& \shortstack{-1.9\\[-1pt]{\scriptsize (1.3)}}
& \shortstack{34.2\\[-1pt]{\scriptsize (4.2)}}
& \shortstack{-59.3\\[-1pt]{\scriptsize (12.3)}}
& \shortstack{4.5\\[-1pt]{\scriptsize (1.4)}}
& \shortstack{-25.5\\[-1pt]{\scriptsize (7.3)}}
& \shortstack{2.8\\[-1pt]{\scriptsize (0.9)}}
& \shortstack{20.1\\[-1pt]{\scriptsize (4.5)}}
& \shortstack{44.7\\[-1pt]{\scriptsize (4.5)}} \\[3pt]

2
& \shortstack{-14.3\\[-1pt]{\scriptsize (6.0)}}
& \shortstack{27.7\\[-1pt]{\scriptsize (3.6)}}
& \shortstack{1.6\\[-1pt]{\scriptsize (0.7)}}
& \shortstack{40.4\\[-1pt]{\scriptsize (4.5)}}
& \shortstack{-1.1\\[-1pt]{\scriptsize (1.0)}}
& \shortstack{36.7\\[-1pt]{\scriptsize (3.9)}}
& \shortstack{-37.9\\[-1pt]{\scriptsize (8.8)}}
& \shortstack{10.6\\[-1pt]{\scriptsize (2.1)}}
& \shortstack{-14.3\\[-1pt]{\scriptsize (5.8)}}
& \shortstack{6.0\\[-1pt]{\scriptsize (1.8)}}
& \shortstack{34.6\\[-1pt]{\scriptsize (4.8)}}
& \shortstack{45.7\\[-1pt]{\scriptsize (4.2)}} \\[3pt]

3
& \shortstack{-2.1\\[-1pt]{\scriptsize (3.1)}}
& \shortstack{33.8\\[-1pt]{\scriptsize (3.9)}}
& \shortstack{2.0\\[-1pt]{\scriptsize (0.8)}}
& \shortstack{43.4\\[-1pt]{\scriptsize (4.2)}}
& \shortstack{0.2\\[-1pt]{\scriptsize (0.9)}}
& \shortstack{38.8\\[-1pt]{\scriptsize (3.9)}}
& \shortstack{-25.4\\[-1pt]{\scriptsize (6.8)}}
& \shortstack{13.9\\[-1pt]{\scriptsize (2.4)}}
& \shortstack{-8.2\\[-1pt]{\scriptsize (5.0)}}
& \shortstack{8.5\\[-1pt]{\scriptsize (2.3)}}
& \shortstack{37.6\\[-1pt]{\scriptsize (4.9)}}
& \shortstack{45.8\\[-1pt]{\scriptsize (4.2)}} \\[3pt]

4
& \shortstack{0.3\\[-1pt]{\scriptsize (3.0)}}
& \shortstack{37.2\\[-1pt]{\scriptsize (3.8)}}
& \shortstack{4.5\\[-1pt]{\scriptsize (1.7)}}
& \shortstack{44.4\\[-1pt]{\scriptsize (4.2)}}
& \shortstack{0.6\\[-1pt]{\scriptsize (0.9)}}
& \shortstack{40.0\\[-1pt]{\scriptsize (4.0)}}
& \shortstack{-15.1\\[-1pt]{\scriptsize (5.0)}}
& \shortstack{16.6\\[-1pt]{\scriptsize (2.5)}}
& \shortstack{-5.3\\[-1pt]{\scriptsize (4.0)}}
& \shortstack{11.2\\[-1pt]{\scriptsize (2.7)}}
& \shortstack{38.4\\[-1pt]{\scriptsize (4.9)}}
& \shortstack{45.8\\[-1pt]{\scriptsize (4.2)}} \\[3pt]

5
& \shortstack{3.0\\[-1pt]{\scriptsize (3.1)}}
& \shortstack{38.4\\[-1pt]{\scriptsize (3.8)}}
& \shortstack{4.7\\[-1pt]{\scriptsize (1.7)}}
& \shortstack{45.3\\[-1pt]{\scriptsize (4.2)}}
& \shortstack{0.7\\[-1pt]{\scriptsize (0.9)}}
& \shortstack{43.2\\[-1pt]{\scriptsize (4.0)}}
& \shortstack{-8.4\\[-1pt]{\scriptsize (3.4)}}
& \shortstack{23.1\\[-1pt]{\scriptsize (3.3)}}
& \shortstack{-4.8\\[-1pt]{\scriptsize (3.9)}}
& \shortstack{14.1\\[-1pt]{\scriptsize (3.3)}}
& \shortstack{38.7\\[-1pt]{\scriptsize (4.9)}}
& \shortstack{45.8\\[-1pt]{\scriptsize (4.2)}} \\[3pt]

6
& \shortstack{5.7\\[-1pt]{\scriptsize (3.4)}}
& \shortstack{39.9\\[-1pt]{\scriptsize (3.9)}}
& \shortstack{6.3\\[-1pt]{\scriptsize (1.9)}}
& \shortstack{45.5\\[-1pt]{\scriptsize (4.2)}}
& \shortstack{0.7\\[-1pt]{\scriptsize (0.9)}}
& \shortstack{43.6\\[-1pt]{\scriptsize (4.1)}}
& \shortstack{-4.7\\[-1pt]{\scriptsize (2.4)}}
& \shortstack{27.3\\[-1pt]{\scriptsize (3.4)}}
& \shortstack{-4.7\\[-1pt]{\scriptsize (3.9)}}
& \shortstack{16.4\\[-1pt]{\scriptsize (3.3)}}
& \shortstack{39.6\\[-1pt]{\scriptsize (4.8)}}
& \shortstack{45.8\\[-1pt]{\scriptsize (4.2)}} \\[3pt]

7
& \shortstack{7.5\\[-1pt]{\scriptsize (3.4)}}
& \shortstack{40.8\\[-1pt]{\scriptsize (3.9)}}
& \shortstack{7.8\\[-1pt]{\scriptsize (1.9)}}
& \shortstack{45.6\\[-1pt]{\scriptsize (4.1)}}
& \shortstack{0.8\\[-1pt]{\scriptsize (0.9)}}
& \shortstack{44.1\\[-1pt]{\scriptsize (4.0)}}
& \shortstack{-4.3\\[-1pt]{\scriptsize (2.4)}}
& \shortstack{29.6\\[-1pt]{\scriptsize (3.6)}}
& \shortstack{-3.4\\[-1pt]{\scriptsize (3.2)}}
& \shortstack{19.4\\[-1pt]{\scriptsize (3.3)}}
& \shortstack{40.3\\[-1pt]{\scriptsize (4.7)}}
& \shortstack{46.1\\[-1pt]{\scriptsize (4.1)}} \\[3pt]

8
& \shortstack{8.5\\[-1pt]{\scriptsize (3.5)}}
& \shortstack{41.5\\[-1pt]{\scriptsize (3.8)}}
& \shortstack{11.2\\[-1pt]{\scriptsize (2.5)}}
& \shortstack{45.8\\[-1pt]{\scriptsize (4.1)}}
& \shortstack{0.9\\[-1pt]{\scriptsize (1.0)}}
& \shortstack{45.1\\[-1pt]{\scriptsize (4.0)}}
& \shortstack{-3.7\\[-1pt]{\scriptsize (2.3)}}
& \shortstack{31.3\\[-1pt]{\scriptsize (3.8)}}
& \shortstack{-2.2\\[-1pt]{\scriptsize (2.4)}}
& \shortstack{20.7\\[-1pt]{\scriptsize (3.5)}}
& \shortstack{40.4\\[-1pt]{\scriptsize (4.7)}}
& \shortstack{46.3\\[-1pt]{\scriptsize (4.0)}} \\[3pt]

9
& \shortstack{9.4\\[-1pt]{\scriptsize (3.6)}}
& \shortstack{41.9\\[-1pt]{\scriptsize (3.8)}}
& \shortstack{12.0\\[-1pt]{\scriptsize (2.5)}}
& \shortstack{45.9\\[-1pt]{\scriptsize (4.1)}}
& \shortstack{0.9\\[-1pt]{\scriptsize (1.0)}}
& \shortstack{45.5\\[-1pt]{\scriptsize (4.0)}}
& \shortstack{-2.9\\[-1pt]{\scriptsize (2.1)}}
& \shortstack{33.0\\[-1pt]{\scriptsize (3.8)}}
& \shortstack{-1.7\\[-1pt]{\scriptsize (2.2)}}
& \shortstack{22.5\\[-1pt]{\scriptsize (3.6)}}
& \shortstack{40.6\\[-1pt]{\scriptsize (4.7)}}
& \shortstack{46.3\\[-1pt]{\scriptsize (4.0)}} \\[3pt]

10
& \shortstack{10.1\\[-1pt]{\scriptsize (3.7)}}
& \shortstack{41.9\\[-1pt]{\scriptsize (3.8)}}
& \shortstack{13.0\\[-1pt]{\scriptsize (2.7)}}
& \shortstack{46.1\\[-1pt]{\scriptsize (4.1)}}
& \shortstack{0.9\\[-1pt]{\scriptsize (1.0)}}
& \shortstack{45.5\\[-1pt]{\scriptsize (4.0)}}
& \shortstack{-2.7\\[-1pt]{\scriptsize (2.1)}}
& \shortstack{33.5\\[-1pt]{\scriptsize (3.9)}}
& \shortstack{-1.4\\[-1pt]{\scriptsize (2.1)}}
& \shortstack{23.1\\[-1pt]{\scriptsize (3.7)}}
& \shortstack{40.9\\[-1pt]{\scriptsize (4.7)}}
& \shortstack{46.3\\[-1pt]{\scriptsize (4.0)}} \\[3pt]

11
& \shortstack{10.5\\[-1pt]{\scriptsize (3.7)}} & --
& \shortstack{13.8\\[-1pt]{\scriptsize (2.9)}} & --
& \shortstack{0.9\\[-1pt]{\scriptsize (1.0)}} & --
& \shortstack{-2.0\\[-1pt]{\scriptsize (1.9)}} & --
& \shortstack{-1.0\\[-1pt]{\scriptsize (2.0)}} & --
& \shortstack{41.2\\[-1pt]{\scriptsize (4.6)}} & -- \\[3pt]

12
& \shortstack{10.8\\[-1pt]{\scriptsize (3.7)}} & --
& \shortstack{14.3\\[-1pt]{\scriptsize (3.0)}} & --
& \shortstack{0.9\\[-1pt]{\scriptsize (1.0)}} & --
& \shortstack{-1.6\\[-1pt]{\scriptsize (1.8)}} & --
& \shortstack{-1.0\\[-1pt]{\scriptsize (2.0)}} & --
& \shortstack{41.4\\[-1pt]{\scriptsize (4.6)}} & -- \\[3pt]

13
& \shortstack{11.6\\[-1pt]{\scriptsize (3.9)}} & --
& \shortstack{14.7\\[-1pt]{\scriptsize (3.0)}} & --
& \shortstack{0.9\\[-1pt]{\scriptsize (1.0)}} & --
& \shortstack{-1.6\\[-1pt]{\scriptsize (1.8)}} & --
& \shortstack{-0.8\\[-1pt]{\scriptsize (1.9)}} & --
& \shortstack{41.7\\[-1pt]{\scriptsize (4.5)}} & -- \\[3pt]

14
& \shortstack{12.1\\[-1pt]{\scriptsize (3.9)}} & --
& \shortstack{14.7\\[-1pt]{\scriptsize (3.0)}} & --
& \shortstack{0.9\\[-1pt]{\scriptsize (1.0)}} & --
& \shortstack{-1.6\\[-1pt]{\scriptsize (1.8)}} & --
& \shortstack{-0.7\\[-1pt]{\scriptsize (1.9)}} & --
& \shortstack{42.0\\[-1pt]{\scriptsize (4.5)}} & -- \\[3pt]

15
& \shortstack{12.2\\[-1pt]{\scriptsize (3.9)}} & --
& \shortstack{15.1\\[-1pt]{\scriptsize (3.0)}} & --
& \shortstack{0.9\\[-1pt]{\scriptsize (1.0)}} & --
& \shortstack{-1.6\\[-1pt]{\scriptsize (1.8)}} & --
& \shortstack{-0.5\\[-1pt]{\scriptsize (1.7)}} & --
& \shortstack{42.1\\[-1pt]{\scriptsize (4.5)}} & -- \\[3pt]

16
& \shortstack{13.2\\[-1pt]{\scriptsize (4.0)}} & --
& \shortstack{15.1\\[-1pt]{\scriptsize (3.0)}} & --
& \shortstack{1.4\\[-1pt]{\scriptsize (0.9)}} & --
& \shortstack{-1.6\\[-1pt]{\scriptsize (1.8)}} & --
& \shortstack{-0.5\\[-1pt]{\scriptsize (1.7)}} & --
& \shortstack{42.1\\[-1pt]{\scriptsize (4.5)}} & -- \\[3pt]

17
& \shortstack{13.8\\[-1pt]{\scriptsize (4.0)}} & --
& \shortstack{15.7\\[-1pt]{\scriptsize (2.9)}} & --
& \shortstack{1.4\\[-1pt]{\scriptsize (0.9)}} & --
& \shortstack{-1.6\\[-1pt]{\scriptsize (1.8)}} & --
& \shortstack{-0.5\\[-1pt]{\scriptsize (1.7)}} & --
& \shortstack{42.1\\[-1pt]{\scriptsize (4.5)}} & -- \\[3pt]

18
& \shortstack{15.3\\[-1pt]{\scriptsize (4.3)}} & --
& \shortstack{15.9\\[-1pt]{\scriptsize (2.9)}} & --
& \shortstack{1.4\\[-1pt]{\scriptsize (0.9)}} & --
& \shortstack{-1.6\\[-1pt]{\scriptsize (1.8)}} & --
& \shortstack{-0.5\\[-1pt]{\scriptsize (1.7)}} & --
& \shortstack{42.2\\[-1pt]{\scriptsize (4.5)}} & -- \\[3pt]

19
& \shortstack{15.8\\[-1pt]{\scriptsize (4.3)}} & --
& \shortstack{16.0\\[-1pt]{\scriptsize (2.9)}} & --
& \shortstack{1.4\\[-1pt]{\scriptsize (0.9)}} & --
& \shortstack{-1.1\\[-1pt]{\scriptsize (1.7)}} & --
& \shortstack{-0.5\\[-1pt]{\scriptsize (1.7)}} & --
& \shortstack{42.2\\[-1pt]{\scriptsize (4.5)}} & -- \\[3pt]

20
& \shortstack{15.9\\[-1pt]{\scriptsize (4.3)}} & --
& \shortstack{16.0\\[-1pt]{\scriptsize (2.9)}} & --
& \shortstack{1.6\\[-1pt]{\scriptsize (0.8)}} & --
& \shortstack{-0.5\\[-1pt]{\scriptsize (1.6)}} & --
& \shortstack{-0.5\\[-1pt]{\scriptsize (1.7)}} & --
& \shortstack{42.5\\[-1pt]{\scriptsize (4.5)}} & -- \\

\bottomrule
\addlinespace[4pt]

\end{tabular}%
}

\begin{minipage}{\textwidth}
\footnotesize
\emph{Notes.} Entries report the mean cost reduction (in \%), rounded to one decimal place, with the standard error shown in parentheses. ``Without'' and ``With'' indicate whether numerical parameter optimization is excluded from or included in the policy-search procedure, respectively. Larger values indicate greater cost reduction. A dash indicates that the corresponding optimizer-enabled experiment was not conducted.
\end{minipage}
\end{table}

\section{LLM Backbone}

This section provides additional results for the LLM-backbone analysis in \S\ref{sec:diff-LLMs}. Table~\ref{tab:llm_backbone_mean_sd} complements Figure~\ref{fig:diff_llm_optimizer_ablation_boxplots} by reporting the generation-by-generation performance of each backbone with and without the external optimizer. The rows correspond to generations \(1\)--\(20\). For each backbone, the two columns report results without and with the optimizer, respectively. Each entry gives the mean cost reduction relative to the optimized base-stock policy, with the corresponding standard error in parentheses. The optimizer-enabled experiments terminate at generation~10, so the corresponding entries are left blank thereafter. The table shows that optimization improves both the level and speed of performance across all six backbones. For example, at generation~10, mean cost reduction with optimization is \(41.9\%\) for GPT-5 Nano, \(46.1\%\) for GPT-5 Mini, \(45.5\%\) for DeepSeek, \(33.5\%\) for Grok 4.1 Fast, \(23.1\%\) for Gemini 2.5 Lite, and \(46.3\%\) for Gemini 3 Flash. Without optimization, even after \(20\) generations, the corresponding values are \(15.9\%\), \(16.0\%\), \(1.6\%\), \(-0.5\%\), \(-0.5\%\), and \(42.5\%\), respectively. The reported standard errors also show substantial heterogeneity across problem instances and repeats, particularly for the stronger-performing backbones.

Table~\ref{tab:llm_backbone_test_convergence} reports the corresponding test-set results at the termination of each search. The columns correspond to the six LLM backbones. The upper block reports the no-optimizer results at generation~20, while the lower block reports the optimizer-enabled results at generation~10. Within each block, the rows report the minimum, maximum, mean, and standard deviation of test-set cost reduction across experimental runs. The test-set results reinforce the in-sample findings. With optimization, mean test-set cost reduction ranges from \(21.2\%\) for Gemini 2.5 Lite to \(45.6\%\) for Gemini 3 Flash. Without optimization, the corresponding means are substantially lower for most backbones, including \(14.8\%\) for GPT-5 Nano, \(15.0\%\) for GPT-5 Mini, \(0.7\%\) for DeepSeek, \(-1.5\%\) for Grok 4.1 Fast, and \(-1.5\%\) for Gemini 2.5 Lite. Gemini 3 Flash remains the main exception, achieving \(41.9\%\) even without optimization. Thus, the performance patterns shown in Figures~\ref{fig:diff_llm_mean_performance} and~\ref{fig:diff_llm_optimizer_ablation_boxplots} persist on the test set.

\begin{table}[t]
\centering
\caption{Test-set cost reduction at convergence across LLM backbones.}
\label{tab:llm_backbone_test_convergence}

\resizebox{\textwidth}{!}{%
\begin{tabular}{llrrrrrr}
\toprule
& 
& GPT-5 Nano
& GPT-5 Mini
& DeepSeek
& Grok 4.1 Fast
& Gemini 2.5 Lite
& Gemini 3 Flash \\
\midrule

\multirow{4}{*}{\shortstack[l]{Without optimizer\\Generation 20}}
& Min
& -81.4 & -1.1 & -4.2 & -24.7 & -49.6 & 9.2 \\
& Max
& 63.8 & 63.7 & 22.7 & 23.7 & 0.7 & 73.1 \\
& Mean
& 14.8 & 15.0 & 0.7 & -1.5 & -1.5 & 41.9 \\
& Std.\ dev.
& 23.9 & 16.6 & 4.4 & 8.3 & 9.1 & 25.2 \\

\addlinespace[4pt]

\multirow{4}{*}{\shortstack[l]{With optimizer\\Generation 10}}
& Min
& 13.0 & 12.5 & 16.1 & 4.3 & -7.2 & 17.3 \\
& Max
& 73.2 & 73.2 & 73.2 & 73.2 & 72.8 & 73.2 \\
& Mean
& 41.3 & 45.4 & 44.9 & 32.8 & 21.2 & 45.6 \\
& Std.\ dev.
& 21.7 & 23.4 & 22.9 & 22.0 & 21.7 & 23.2 \\
\bottomrule
\addlinespace[4pt]

\end{tabular}\vspace{6pt}
}
\begin{minipage}{\textwidth}
\footnotesize
\emph{Notes.} Entries summarize the test-set cost reduction (in \%) across experimental runs at convergence. The without-optimizer results correspond to generation 20, whereas the with-optimizer results correspond to generation 10. Larger values indicate greater cost reduction.
\end{minipage}
\end{table}
\end{document}